\documentclass[10pt,twocolumn,letterpaper]{article}

\usepackage{iccv}
\usepackage[dvipsnames, svgnames, x11names]{xcolor}
\usepackage{times}
\usepackage{epsfig}
\usepackage{graphicx}
\usepackage{subcaption}
\usepackage{amsmath}
\usepackage{amssymb}
\usepackage{adjustbox}
\usepackage{bm}
\usepackage[ruled,vlined]{algorithm2e}
\usepackage{multirow}
\usepackage{booktabs}
\usepackage{bbding}
\usepackage{multirow,booktabs}
\usepackage{colortbl}
\usepackage{booktabs}
\PassOptionsToPackage{hyphens}{url}
\usepackage[pagebackref=true,breaklinks=true,letterpaper=true,colorlinks,bookmarks=false]{hyperref}
\usepackage[capitalize]{cleveref}

\crefname{section}{Sec.}{Secs.}
\Crefname{section}{Section}{Sections}
\Crefname{table}{Table}{Tables}
\crefname{table}{Tab.}{Tabs.}

\usepackage[breaklinks=true,bookmarks=false]{hyperref}

\iccvfinalcopy 

\def\iccvPaperID{****} 

\ificcvfinal\fi

\begin{document}

\title{HumanMAC: Masked Motion Completion for Human Motion Prediction
\vspace{-1.5em}
}

\author{
\vspace{0.3em}
\url{https://lhchen.top/Human-MAC}\\
Ling-Hao Chen$^{1}$\thanks{This project is led by Ling-Hao Chen and Xiaobo Xia jointly.} \thanks{Equal contribution. }  \ \ \ \ Jiawei Zhang$^{2}$$^{\dag}$ \ \ \ \ Yewen Li$^{3}$ \ \ \ \ Yiren Pang$^{2}$ \ \ \ \ Xiaobo Xia$^{4}$$^{*}$\thanks{Corresponding author.} \ \ \ \ Tongliang Liu$^{4}$ \\
$^1$Tsinghua University \quad\quad
$^2$Xidian University \\
$^3$Nanyang Technological  University\quad\quad
$^4$The University of Sydney\\
\small
\vspace{-0.4em} 
\texttt{\{thu.lhchen\}@gmail.com} \ \ \ \ \ 
\texttt{\{zjw\}@stu.xidian.edu.cn} \ \ \ \ \  
\texttt{\{yewen001\}@e.ntu.edu.sg} \\ 
\small
\vspace{-2.5em} 
\texttt{\{yrpang\}@outlook.com} \ \ \ \ \ 
\texttt{\{xiaoboxia.uni\}@gmail.com} \ \ \ \ \  
\texttt{\{tongliang.liu\}@sydney.edu.au}
}
\maketitle

\ificcvfinal\thispagestyle{empty}\fi

\newcommand{\myPara}[1]{\vspace{.05in}\noindent\textbf{#1}}
\begin{abstract}
\vspace{-0.4em} 
    Human motion prediction is a classical problem in computer vision and computer graphics, which has a wide range of practical applications. Previous effects achieve great empirical performance based on an encoding-decoding style. The methods of this style work by first encoding previous motions to latent representations and then decoding the latent representations into predicted motions. However, in practice, they are still unsatisfactory due to several issues, including complicated loss constraints, cumbersome training processes, and scarce switch of different categories of motions in prediction. In this paper, to address the above issues, we jump out of the foregoing style and propose a novel framework from a new perspective. Specifically, our framework works in a masked completion fashion. In the training stage, we learn a motion diffusion model that generates motions from random noise. In the inference stage, with a denoising procedure, we make motion prediction conditioning on observed motions to output more continuous and controllable predictions. The proposed framework enjoys promising algorithmic properties, which only needs one loss in optimization and is trained in an end-to-end manner. Additionally, it accomplishes the switch of different categories of motions effectively, which is significant in realistic tasks, \textit{e.g.}, the animation task. Comprehensive experiments on benchmarks confirm the superiority of the proposed framework. The project page is available at \url{https://lhchen.top/Human-MAC}. 
    \vspace{-1.7em}
\end{abstract}

\section{Introduction}
\label{sec:intro}

\begin{figure}[!t]
\centering
    \hspace{-21pt}
    \begin{subfigure}{0.43\linewidth}
        \centering
        \includegraphics[width=1.0\linewidth]{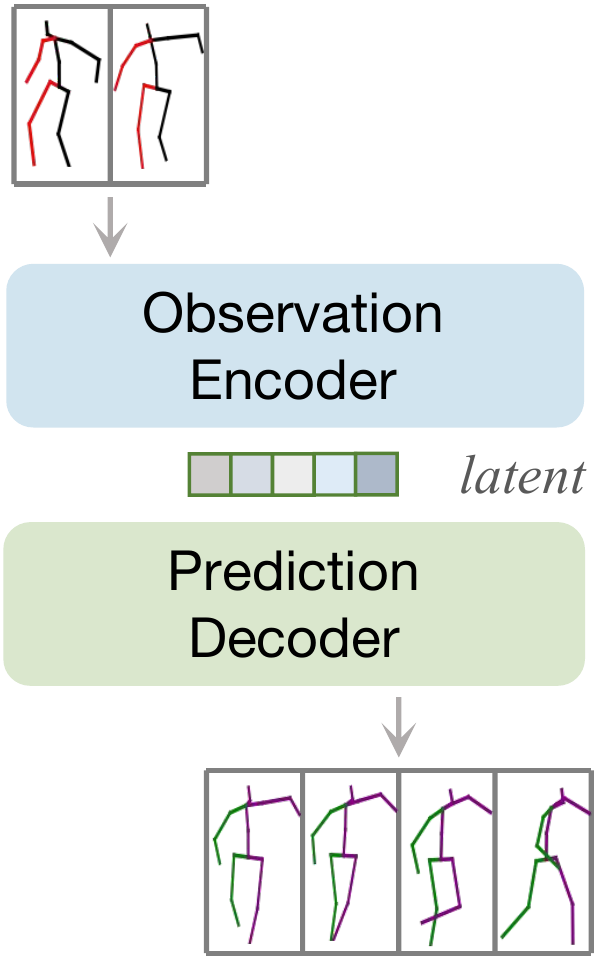}
        \caption{\footnotesize{encoding-decoding}}
        \label{fig:insighta}
    \end{subfigure}
    \hspace{5pt}
\centering
    \begin{subfigure}{0.43\linewidth}
        \centering
        \includegraphics[width=1.0\linewidth]{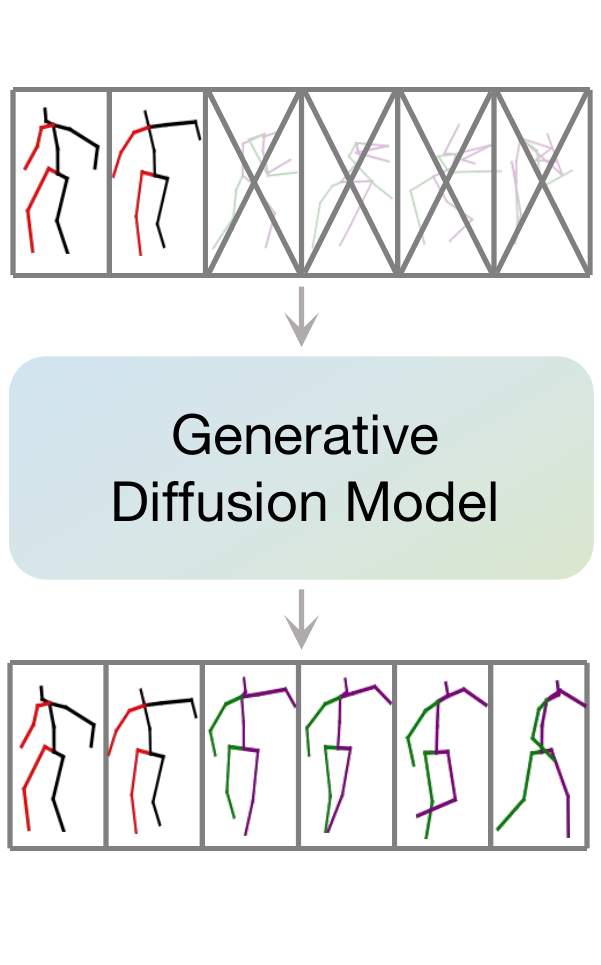}
        \caption{\footnotesize{masked completion (\textit{\textbf{Ours}})}}
        \label{fig:insightb}
    \end{subfigure}
\centering
    \begin{subfigure}{0.88\linewidth}
        \centering
        \includegraphics[width=1\linewidth]{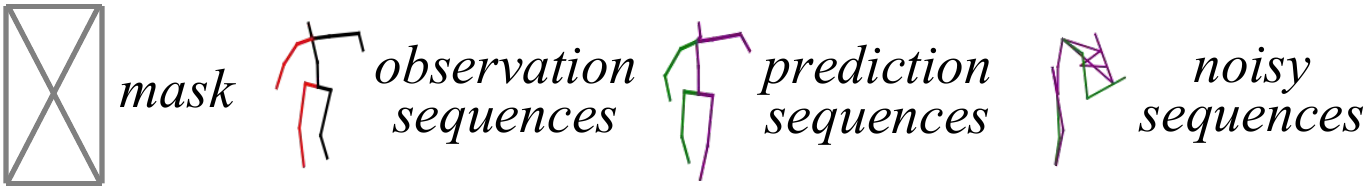}
        \label{fig:insightc}
    \end{subfigure}
\vspace{-5.5mm}
 \caption{Comparison between the encoding-decoding fashion and masked motion completion. (a) The methods in the encoding-decoding fashion encode the observation into a latent explicitly and then decodes the latent into prediction results. (b) The proposed motion diffusion model generates motions from noise in the training stage. In the inference stage, it treats HMP as a masked completion task.}
 \vspace{-6.5mm}
\end{figure}

The problem of Human Motion Prediction~(HMP) focuses on predicting possible future pose sequences from a sequence of observed motions~\cite{barsoum2018hp, fragkiadaki2015recurrent, bhattacharyya2018accurate, gurumurthy2017deligan, zhang2019predicting, yan2021dms}, which has a wide range of applications~\cite{yan2018mt, ju2023humanart, yuan2019diverse, salzmann2022motron, mao2021generating, yuan2020dlow, yang2023nural, ju2023humansd, zhu2022bars}, \textit{e.g.}, autonomous driving~\cite{paden2016survey} and healthcare~\cite{troje2002decomposing}. This problem is complicated and challenging since the predicted motions are always required to be \textit{continuous}, \textit{diverse}, and \textit{realistic} simultaneously~\cite{dang2022diverse}. 

Prior state-of-the-art methods work in an encoding-decoding fashion to tackle the HMP problem, which conditions on previous motion frames and predicts unobserved motions~\cite{salzmann2022motron, lucas2022posegpt, blattmann2021behavior, dang2022diverse, xu22stars}. Technically, these methods first encode the previous motion frames to latent representations explicitly and then decode the latent representations into prediction results~(see Figure~\ref{fig:insighta})~\cite{petrovich2021action, wei2022human, barquero2022belfusion, hassan2021stochastic, adeli2021tripod}. 

Although these methods enjoy the good performance in some scenarios, they are still unsatisfactory in practice. We detail the issues from three aspects. (1) Most state-of-the-art methods rely on multiple loss constraints for high-quality prediction results, \textit{e.g.}, the average pairwise distance~\cite{aliakbarian2020stochastic}, final displacement error~\cite{gupta2018social, lee2017desire}, and adversarial loss~\cite{barsoum2018hp, gurumurthy2017deligan}. Consequently, they need carefully designed hyper-parameters to balance different loss constraints, which makes it laborious for method applications. (2) Previous state-of-the-art methods need multi-stage training~\cite{yuan2020dlow, barquero2022belfusion}. That is to say, the learning of the encoder/decoder and sampling in the latent space is performed in different stages. To make matters worse, complex pipelines always require an additional stage of engineering tuning. (3) For the methods, it is hard to realize the
switch of different categories of motions, \textit{e.g.}, switch from \texttt{WalkDog} to \texttt{Sitting}, which is pivotal for result diversity. The reason is that these methods are largely limited to observed motion sequences for prediction, which include few such switches.

In this paper, we get out of the encoding-decoding fashion and propose the \textbf{MA}sked motion \textbf{C}ompletion framework to handle the HMP problem~(\textit{aka} \textbf{HumanMAC}). To the best of our knowledge, this is the first time to tackle HMP from the perspective of masked completion. Specifically, 
in the training stage, we integrate observed and predicted motions, and learn a motion diffusion model to generate motions from random noise. In the inference stage, we predict possible future motions given observed motions. Note that traditional diffusion models~\cite{ho2020denoising, song2020denoising, ludpm, nichol2021improved, lu2022dpm, yang2023boosting} have to make the prediction from random noise with a denoising procedure, which cannot make use of observed motions, resulting in uncontrollable results. We thereby propose a new method named DCT-Completion in the inference stage, which can employ observed motions for future motion prediction. Particularly, we add noise to observations to obtain a noisy spectrum of observed motions. In each denoising step of our method, we involve this noisy spectrum with a masking mechanism, where we combine the denoised and noisy spectrum with a mask. As the noisy spectrum is transformed from observations, the mechanism makes prediction motions conditioned on observed motions. Controllable predictions are produced accordingly~(see Figure~\ref{fig:insightb}).

Compared with the methods in the encoding-decoding fashion, HumanMAC enjoys the following great properties. (1) During training, there is only one loss function in the objective of HumanMAC. This avoids previous several hyper-parameters that balance different losses, and facilitates method applications. (2) HumanMAC is trained in an end-to-end manner, which is remarkably simpler to implement than multi-stage training. (3) HumanMAC can achieve more diverse prediction results that contain the switch of different categories of motions, even though the switch is rare or not presented in observed motion sequences. This is benefited from that we holistically model the whole sequence, \textit{i.e.}, the integrated observed and predicted motions. The generated motion
is therefore controlled to be continuous between the observed frames
and predicted frames. Due to the continuity of motions, \textit{e.g.}, from \texttt{Squatting} to \texttt{Standing Up} and from \texttt{Standing Still} to \texttt{Walking} for we humans, the trained model can naturally complete the switch of different categories of motions.

Before delving into details, we clearly summarize our contributions as follows:
\begin{itemize}
    \item We carefully discuss the three issues of previous paradigms in human motion prediction, which can inspire follow-up research for algorithm improvement. 
    \item To address the issues, we propose a novel framework from the new perspective of masked completion. The proposed method enjoys promising algorithmic properties that include only one loss function in the objective, an end-to-end training manner, and great motion switch ability. 
    \item We conduct a series of experiments on benchmark datasets to justify our claims. In both qualitative and visualization comparisons with the state-of-the-art methods, our method achieves superior performance, which creates a simple and strong baseline for future research. Comprehensive ablation studies and discussions are also provided. 
\end{itemize}

\section{Related Work}

\subsection{Human Motion Prediction}
Early in the research, traditional works~\cite{ijcai2022p111, martinez2017human, li2018convolutional, tang2018long, corona2020context, hernandez2019human, mao2019learning, sofianos2021space, zhong2022spatio, sofianos2021space} try to predict motions in a deterministic way. In consideration that the prediction of motions is subjective, a series of works propose to produce diverse motions in prediction. At the present stages, the state-of-the-art methods of human motion prediction predict motions in an encoding-decoding way~\cite{salzmann2022motron, lucas2022posegpt, blattmann2021behavior, dang2022diverse, xu22stars}, and rely on carefully designed multiple loss constraints~\cite{aliakbarian2020stochastic, gupta2018social, lee2017desire} to achieve the diversity and authenticity of human motions. Besides, most of the previous state-of-the-art methods need multi-stage training and cannot be trained in an end-to-end way. This property results in its dependency on the pre-training quality of the encoder and decoder. Therefore, we propose a human motion diffusion model with only one loss, which can be trained in an end-to-end way. Furthermore, thanks to the advantages of the completion method in the inference stage, we can switch between different motions for generating richer human motions, which is not achieved by previous works.

\subsection{Denoising Diffusion Models}

Denoising diffusion probabilistic models~\cite{ho2020denoising, nichol2021improved, song2020denoising, lu2022dpm,chen2022diffusiondet,wang2023diffusion, sun2023corrmatch} are motivated by the second law of thermodynamics. The models try to learn a reverse diffusion process from random noise to the data distribution. Due to its dynamics, the generated results are both diverse and high-quality~\cite{pearce2023imitating,xiao2021tackling,whang2022deblurring}. Therefore, diffusion models have also produced great impacts on image/video generation~\cite{dhariwal2021diffusion, rombach2022high, yu2022thpad, hovideo}, drug discovery~\cite{xugeodiff}, 3D reconstruction~\cite{xu2022dream3d}, and other research problems~\cite{popov2021grad, yang2023semantic, nichol2022glide, kongdiffwave, han2022card, lugmayr2022repaint, zhang2023t2m}. To take advantage of its good properties, we introduce the denoising diffusion model to predict human motions.

Before our work, MotionDiffuse~\cite{zhang2022motiondiffuse} and MDM~\cite{tevet2022human} are prior works that introduce diffusion models into the text-driven motion synthesis area. Tseng \textit{et. al.} propose EDGE~\cite{tseng2022edge} to generate motions from music. Alexanderson \textit{et. al.} propose a diffusion-based model~\cite{alexanderson2022listen} to synthesize motion driven by the audio. UDE~\cite{zhou2022ude} and MoFusion~\cite{dabral2022mofusion} propose to unify the audio-/text-driven motion generation task~\cite{li2021ai, petrovich2022temos, hong2022avatarclip, tevet2022motionclip, siyao2022bailando, chen2022mld, petrovich2021action, plappert2016kit, li2022danceformer, jiang2023motiongpt} in one system. BeLFusion~\cite{barquero2022belfusion} tries to predict motions via the diffusion model in the latent space. However, it disentangles the model training into several stages, which is restricted by the pre-training quality of the encoder and decoder. Our end-to-end training framework can avoid this problem.

\section{Preliminaries}
\subsection{Problem Formulation}
We note the observed sequence of the $H$-frame motion as $\mathbf{x}^{\left( 1:H \right)}=\left[ \mathbf{x}^{\left( 1 \right)};\mathbf{x}^{\left( 2 \right)};\ldots ; \mathbf{x}^{\left( H \right)} \right] \in \mathbb{R} ^{H\times 3J}$, where $\mathbf{x}^{\left(h \right)}\in \mathbb{R} ^{3J}$ is the coordinates of each joints at the frame $h$ and $J$ is the number of joints. Given the observed motion $\mathbf{x}^{\left( 1:H \right)}$, the objective of the Human Motion Prediction~(HMP) problem is to predict the following $F$ motions $\mathbf{x}^{\left( H+1: H+F \right)}=\left[ \mathbf{x}^{\left( H+1 \right)};\mathbf{x}^{\left( H+2 \right)};\ldots;\mathbf{x}^{\left( H+F \right)} \right] \in \mathbb{R} ^{F\times 3J}$.

\subsection{Discrete Cosine Transform}
\label{sec:dct}

The work~\cite{mao2019learning} proposed the Discrete Cosine Transform (DCT) to predict and generate human motions. The DCT operation extracts both current and periodic temporal properties from the motion sequence, which is beneficial for obtaining continuous motions. Therefore, staying precedent~\cite{mao2019learning}, we train our model with DCT.

Technically, given a ($H$+$F$)-frame motion sequence $\mathbf{x}\in \mathbb{R} ^{\left(H+F\right) \times 3J}$, we project the sequence into the DCT domain via the $\texttt{DCT}(\cdot)$ operation:
\begin{equation}
    \mathbf{y} = \texttt{DCT}(\mathbf{x}) = \boldsymbol{D}\mathbf{x},
    \label{eq:dct}
\end{equation}
where $\boldsymbol{D}\in \mathbb{R} ^{\left(H+F\right) \times \left(H+F\right)}$ is the predefined DCT basis~\cite{mao2019learning}, and $\mathbf{y}\in \mathbb{R} ^{\left(H+F\right) \times 3J}$ is the DCT coefficients. As the DCT operation is an orthogonal transform, we can recover the motion sequence from the DCT domain via an inversed Discrete Cosine Transform ($\texttt{iDCT}(\cdot)$) operation:
\begin{equation}
    \mathbf{x} = \texttt{iDCT}(\mathbf{y}) = \boldsymbol{D}^\top\mathbf{y}.
    \label{eq:idct}
\end{equation}

Due to the smoothness property of human motions~\cite{mao2019learning}, we simplify to perform the DCT and iDCT operation by selecting the first $L$ rows of $\boldsymbol{D}$ and $\boldsymbol{D}^\top$ (noted as $\boldsymbol{D}_L,\boldsymbol{D}_L^\top\in \mathbb{R}^{L\times (H+F)}$), following $\mathbf{c}\approx\boldsymbol{D}_L\mathbf{x}$ and $\mathbf{x}\approx\boldsymbol{D}_L^\top\mathbf{c}$. The simplified form of the DCT/iDCT operation can reduce the computational cost by discarding the high-frequency components~\cite{mao2019learning}. In the sequel, if there is no confusion, we will replace $\boldsymbol{D}$ with $\boldsymbol{D}_L$.

\vspace{-0.8em}
\section{Methodology}
\vspace{-2pt}

\subsection{Model Training}
\vspace{-2pt}
In the training stage, we conduct the DCT operation discussed above on the full motion $\mathbf{x}\in\mathbb{R} ^{\left( H+F \right) \times 3J}$ for the first $L$ frequency components to obtain the spectrum $\mathbf{y}_0\in \mathbb{R} ^{L\times 3J}$, where $\mathbf{y}_0=\mathbf{y}$ in Eq.~(\ref{eq:dct}). The noisy DCT spectrum $\mathbf{y}_t$ at the timestep $t$ can be calculated by the reparameterization trick:
\begin{equation}
\mathbf{y}_t=\sqrt{\bar{\alpha}_t}\mathbf{y}_0+\sqrt{1-\bar{\alpha}_t}\boldsymbol{\epsilon}, 
\end{equation}
where $\bar{\alpha}_{t}=\prod_{i=1}^{t} \alpha_{i}$, $\alpha_{i}\in [0, 1]$ are pre-defined variance parameters, and $\boldsymbol{\epsilon} \sim \mathcal{N} \left( \mathbf{0}, \mathbf{I} \right)$. There are $T$ timesteps in total. In this paper, for noise prediction, we employ a simple network parameterized by $\bm{\theta}$, whose architecture is shown in Appendix~\ref{sec:network}. We name the noise prediction network as \texttt{TransLinear}. At the timestep $t$, the network outputs predicted noise as $\bm{\epsilon}_{\bm{\theta}}\left( \mathbf{y}_t, t \right)$. Afterward, we optimize the parameters $\bm{\theta}$ with the noise prediction loss:
\begin{equation}\label{eq:noise_prediction}
    \mathcal{L} =\mathbb{E} _{\bm{\epsilon} ,t}\left[ \left\| \boldsymbol{\epsilon} - \boldsymbol{\epsilon}_{\bm{\theta}}\left( \mathbf{y}_t,t \right) \right\| ^2 \right].
\end{equation}
It is worth noting that the loss in Eq.~(\ref{eq:noise_prediction}) is the \textit{only loss} in our HumanMAC during training. Additionally, HumanMAC is trained in an end-to-end manner by minimizing the loss $\mathcal{L}$ directly. We provide an algorithm flow of model training in Algorithm~\ref{alg:train}.

\begin{figure}[!t]
	\centering
        \hspace{-20pt}
	\begin{subfigure}{0.38\linewidth}
		\centering
  \includegraphics[width=1.1\linewidth]{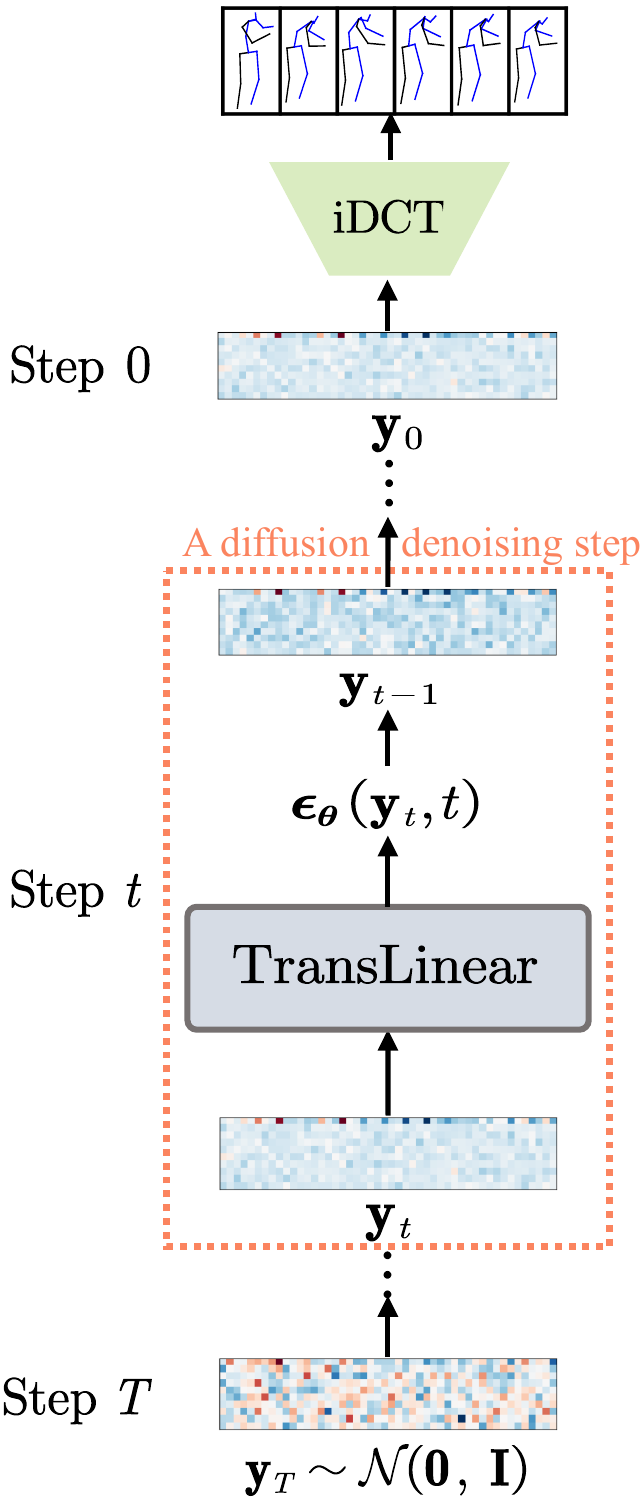}
		\caption{Prior diffusion generation with $T$ steps.}
		\label{diff}
	\end{subfigure}
    \hspace{10pt}
	\centering
	\begin{subfigure}{0.38\linewidth}
		\centering
		\includegraphics[width=1.2\linewidth]{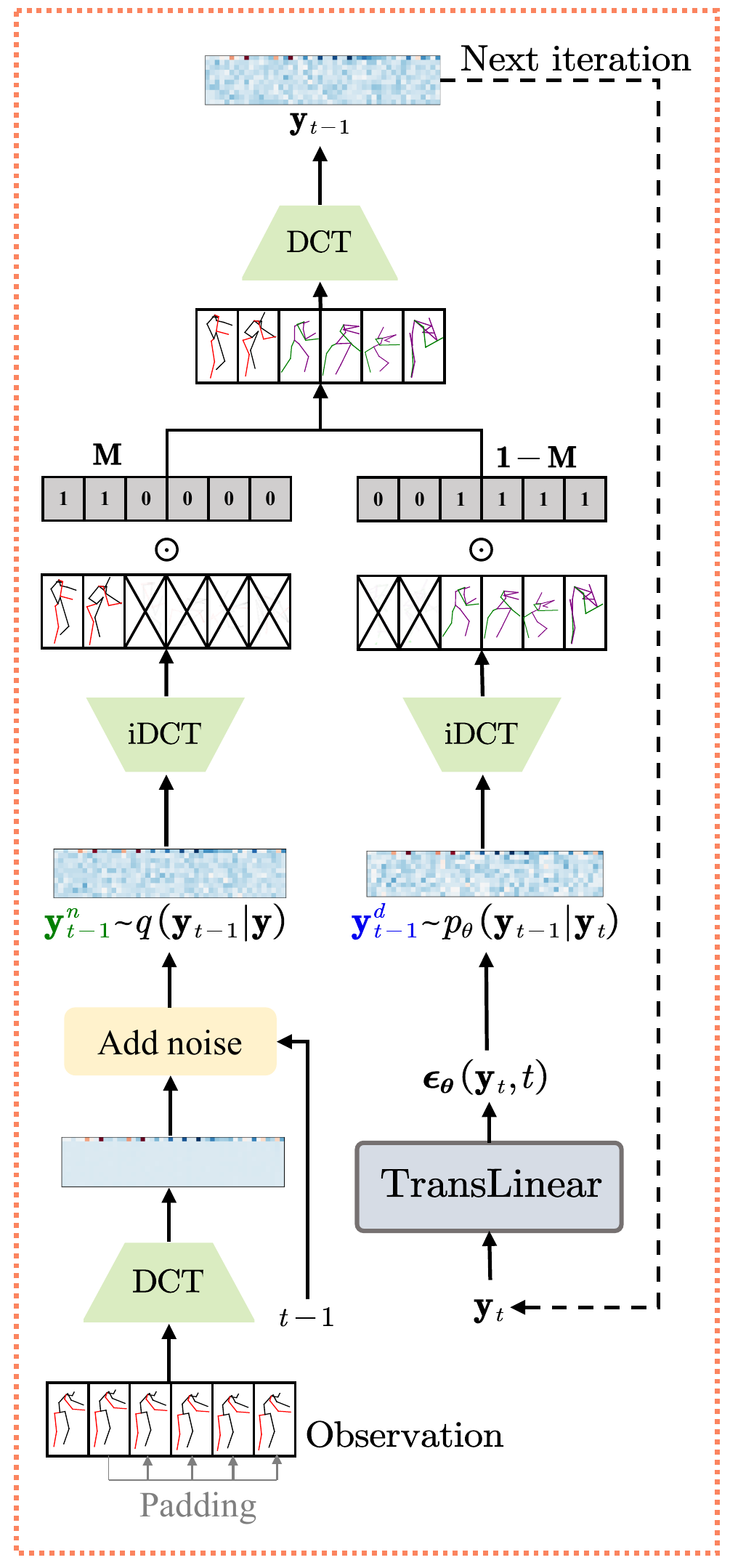}
		\caption{A diffusion DCT-Completion step.}
		\label{dctc}
	\end{subfigure}
 \vspace{-3.5mm}
 \caption{Comparison between the vanilla denoising procedure and the proposed DCT-Completion.}
 \vspace{-10pt}
\end{figure}

\vspace{-0.5em}
\subsection{DCT-Completion in Inference}
\label{sec:paint}
In the inference stage, given observed motions, we need to use the trained model to make predicted motions. In prior diffusion-based works, with a trained diffusion model, motions can be generated via $T$ denoising steps. We provide the illustration of this generation in Figure~\ref{diff}. However, the prior procedure cannot make use of observed motions, which makes predicted motions  uncontrollable.  
To address the issue, we propose a new completion algorithm without model retraining, named DCT-Completion, which can complete prediction motions with observation motions. We discuss DCT-Completion as follows.

\myPara{Add noise into observation.} 
For the left branch in Figure~\ref{dctc}, we pad the observation into a full sequence with the final observation frame and project it to the DCT domain, noted as $\mathbf{y}$. Then, we add noise on $\mathbf{y}$ to obtain noisy spectrum of the observation at the timestep $t-1$: 
\begin{equation}
    \mathbf{y}_{t-1}^{n}=\sqrt{\bar{\alpha}_{t-1}}\mathbf{y}+\sqrt{1-\bar{\alpha}_{t-1}}\mathbf{z} ,
    \label{eq:deoise}
\end{equation}
where $\mathbf{z}\sim \mathcal{N}(\mathbf{0}, \mathbf{I})$. Note that $\mathbf{y}_{t-1}^{n} \sim q(\mathbf{y}_{t-1} \mid \mathbf{y})$. 

\myPara{Denoised prediction.} For the right denoising branch shown in Figure~\ref{dctc}, it aims to denoise the noisy prediction spectrum $\mathbf{y}_{t-1}$ from $\mathbf{y}_{t}$: 
\begin{equation}
     \mathbf{y}_{t-1}^{d} = \frac{1}{\sqrt{\alpha_{t}}}\left(\mathbf{y}_{t}-\frac{1-\alpha_{t}}{\sqrt{1-\bar{\alpha}_{t}}} \boldsymbol{\epsilon}_{\bm{\theta}}\left(\mathbf{y}_{t}, t\right)\right) +\sigma_{t} \mathbf{z},
     \label{eq:noise}
\end{equation}
where $\mathbf{z}\sim \mathcal{N}(\mathbf{0}, \mathbf{I})$ if $t=1$; else $\mathbf{z}=\mathbf{0}$. Note that $\mathbf{y}_{t-1}^{d} \sim p_{\bm{\theta}}(\mathbf{y}_{t-1}\mid \mathbf{y}_{t})$. 

\myPara{Masked completion.} In a trained diffusion model, the noisy observation spectrum $\mathbf{y}_{t-1}^{n}$ and denoised prediction spectrum $\mathbf{y}_{t-1}^{d}$ are approximately the same distribution~\cite{ho2020denoising}, \textit{i.e.}, $q(\mathbf{y}_{t-1} \mid \mathbf{y}) \approx p_{\bm{\theta}}(\mathbf{y}_{t-1}\mid \mathbf{y}_{t})$. Therefore, we project the denoised and noisy spectrum to the temporal domain via the iDCT operation and then combine them together with a masking mechanism:
\begin{equation}
    \label{eq:paint}
    \begin{aligned}
        {\small
        \mathbf{y}_{t-1}  = \texttt{DCT}[\mathbf{ M}\odot \texttt{iDCT}( \mathbf{y}_{t-1}^{n})+(\mathbf{1}-\mathbf{M}) \odot \texttt{iDCT}(\mathbf{y}_{t-1}^{d})],
        }
    \end{aligned}
\end{equation}
where $\mathbf{M} = [\underbrace{1, 1, \ldots, 1}_{H-\text{dim}}, \underbrace{0, 0, \ldots 0}_{F-\text{dim}}]^\top$ denotes the mask of the prediction, and $\odot $ denotes the Hadamard product. With the masked noisy observation guidance, we can complete the prediction motion in each reverse diffusion step. 

For a convenient understanding of readers, we provide the algorithm flow of inference in Algorithm~\ref{alg:paint}. 

\myPara{Motion switch.} For the motion switch ability, different from the vanilla HMP task, we  provide the final additional $M$-frame target motions in the inference stage to obtain the controllable prediction. Besides, we will mask the motions except for observation and target motions, \textit{i.e.}, $\mathbf{M}= [\underbrace{1, 1, \ldots, 1}_{H-\text{dim}}, \underbrace{0, 0, \ldots 0}_{(F-M)-\text{dim}}, \underbrace{1, 1, \ldots 1}_{M-\text{dim}}]^\top$. As we model the whole sequence directly, the generated motion is controlled to be continuous among the observed, predicted, and target frames. Besides, benefiting from the continuity of motions, the trained model can naturally complete the switch of different categories of motions, \textit{e.g.}, from \texttt{Sitting} to \texttt{Standing}.

\begin{algorithm}[!t]
    \caption{Training procedure of HumanMAC}
    \KwIn{motion $\mathbf{x}\in \mathbb{R} ^{\left(H+F\right) \times 3J}$, noising steps $T$, the initialized noise prediction network $\boldsymbol{\epsilon}_{\bm{\theta}}$, maximum iterations $I_{\text{max}}$.}
    \KwOut{the noise prediction network $\boldsymbol{\epsilon}_{\bm{\theta}}$.}

    \For{$I=0, 1,\ldots, I_{\text{\normalfont{max}}}$}{
        $\mathbf{y}_0 = \texttt{DCT}(\mathbf{x}) \sim p(\mathbf{y}_0)$\;
        $t = \texttt{Uniform}(\{1, 2, \cdots, T\})$\;
        $\boldsymbol{\epsilon} \sim \mathcal{N}(\mathbf{0}, \mathbf{I})$\;
        $\bm{\theta} = \bm{\theta} - \nabla_{\bm{\theta}} \left\|\boldsymbol{\epsilon}-\boldsymbol{\epsilon}_{\bm{\theta}}\left(\sqrt{\bar{\alpha}_{t}} \mathbf{y}_{0}+\sqrt{1-\bar{\alpha}_{t}} \boldsymbol{\epsilon}, t\right)\right\|^{2}$\;
    }
    \textbf{return} the noise prediction network $\boldsymbol{\epsilon}_{\bm{\theta}}$.
    \label{alg:train}
\end{algorithm}
\vspace{-5pt}
\begin{algorithm}[!t]
    \caption{Inference procedure of HumanMAC}
    \KwIn{observed motion $\mathbf{x}^{(1: H)}\in \mathbb{R} ^{H \times 3J}$, the mask of the observation $\mathbf{M}$, noising steps $T$, the trained noise prediction network $\boldsymbol{\epsilon}_{\bm{\theta}}$.}
    \KwOut{competed motion $\mathbf{x} \in \mathbb{R} ^{\left(H+F\right) \times 3J}$.}
    $\mathbf{y}_{T} \sim \mathcal{N}(\mathbf{0}, \mathbf{I})$\;
    $\mathbf{x} := \texttt{Pad}(\textbf{x})\in \mathbb{R} ^{\left(H+F\right) \times 3J}$ \ // observation padding\;
    $\mathbf{y} = \texttt{DCT}(\mathbf{x}) \sim p(\mathbf{y})$\;
    \For{$t \in T, T-1,..., 1$}{
        $\mathbf{z} \sim \mathcal{N}(\mathbf{0}, \mathbf{I})$ if $t>1$, else $\mathbf{z} = \textbf{0}$\;
        ${\color{Green} \mathbf{y}_{t-1}^{n}} =  \sqrt{\bar{\alpha}_{t-1}}\mathbf{y}+\sqrt{1-\bar{\alpha}_{t-1}}\boldsymbol{z}$\;
        ${{\color{Blue} \mathbf{y}_{t-1}^{d}} = \frac{1}{\sqrt{\alpha_{t}}}\left(\mathbf{y}_{t}-\frac{1-\alpha_{t}}{\sqrt{1-\bar{\alpha}_{t}}} \boldsymbol{\epsilon}_{\bm{\theta}}\left(\mathbf{y}_{t}, t\right)\right)+\sigma_{t} \mathbf{z}}$\;
        $\mathbf{y}_{t-1}  = \texttt{DCT}[\mathbf{M}\odot\texttt{iDCT}( {\color{Green} \mathbf{y}_{t-1}^{n}}) + \texttt{iDCT}((\mathbf{1}-\mathbf{M}) \odot {\color{Blue} \mathbf{y}_{t-1}^{d}}])$\;
    }
    \textbf{return} $\texttt{iDCT}(\mathbf{y}_0)$.
    \label{alg:paint}
\end{algorithm}

\begin{center}
\begin{table*}[!t]
\small
\centering
\setlength\tabcolsep{1.5pt}
\begin{tabular}{ccc|ccccc|ccccc}
\toprule
\textbf{}        & \multicolumn{1}{c}{\multirow{2}{*}{\small\begin{tabular}[c]{@{}c@{}}{\makebox[0.07\textwidth][c]{One-Stage}} \end{tabular}}} & \multirow{2}{*}{\small\begin{tabular}[c]{@{}c@{}}{\# Loss}\end{tabular}} & \multicolumn{5}{c|}{Human3.6M}         & \multicolumn{5}{c}{HumanEva-I}        \\ \cline{4-13} 
                 & \multicolumn{1}{c}{}                      &                        & \makebox[0.06\textwidth][c]{APD$\uparrow$}   & \makebox[0.065\textwidth][c]{ADE$\downarrow$}    & \makebox[0.065\textwidth][c]{FDE$\downarrow$}   & \makebox[0.075\textwidth][c]{MMADE$\downarrow$} & \makebox[0.075\textwidth][c]{MMFDE$\downarrow$} & \makebox[0.06\textwidth][c]{APD$\uparrow$}   & \makebox[0.065\textwidth][c]{ADE$\downarrow$}   & \makebox[0.065\textwidth][c]{FDE$\downarrow$}   & \makebox[0.075\textwidth][c]{MMADE$\downarrow$} & \makebox[0.075\textwidth][c]{MMFDE$\downarrow$} \\ \hline
acLSTM~\cite{zhou2018autoconditioned}           & {\Checkmark}                                         & \multicolumn{1}{c|}{1} & 0.000  & 0.789 & 1.126 & 0.849 & 1.139 & 0.000 & 0.429 & 0.541 & 0.530 & 0.608 \\
DeLi GAN~\cite{gurumurthy2017deligan}         & {\Checkmark}                                         & \multicolumn{1}{c|}{1} & 6.509  & 0.483 & 0.534 & 0.520 & 0.545 & 2.177 & 0.306 & 0.322 & 0.385 & 0.371 \\
MT-VAE~\cite{yan2018mt}           & {\Checkmark}                                         & \multicolumn{1}{c|}{3} & 0.403  & 0.457 & 0.595 & 0.716 & 0.883 & 0.021 & 0.345 & 0.403 & 0.518 & 0.577 \\
BoM~\cite{bhattacharyya2018accurate}              & {\Checkmark}                                         & \multicolumn{1}{c|}{1} & 6.265  & 0.448 & 0.533 & 0.514 & 0.544 & 2.846 & 0.271 & 0.279 & 0.373 & 0.351 \\
DSF~\cite{yuan2019diverse}              & {\XSolidBrush}                                         & \multicolumn{1}{c|}{2} & 9.330  & 0.493 & 0.592 & 0.550 & 0.599 & 4.538 & 0.273 & 0.290 & 0.364 & 0.340 \\
DLow\cite{yuan2020dlow}             & {\XSolidBrush}                                         & \multicolumn{1}{c|}{3} & 11.741 & 0.425 & 0.518 & 0.495 & 0.531 & 4.855 & 0.233 & 0.244 & 0.343 & 0.331 \\
GSPS~\cite{mao2021generating}             & {\XSolidBrush}                                           & \multicolumn{1}{c|}{5} & 14.757 & 0.389 & 0.496 & 0.476 & 0.525 & 5.825 & 0.233 & 0.244 & 0.343 & 0.331 \\
MOJO~\cite{zhang2021we}             & {\XSolidBrush}                                           & \multicolumn{1}{c|}{3} & 12.579 & 0.412 & 0.514 & 0.497 & 0.538 & 4.181 & 0.234 & 0.244 & 0.369 & 0.347 \\
BeLFusion~\cite{barquero2022belfusion}        & {\XSolidBrush}                                          & \multicolumn{1}{c|}{4} & 7.602  & 0.372 & \textbf{0.474} & \textbf{0.473} & \textbf{0.507} & -     & -     & -     & -     & -     \\
DivSamp~\cite{dang2022diverse}          & {\XSolidBrush}                                           & \multicolumn{1}{c|}{3} & 15.310 & 0.370 & 0.485 & 0.475 & 0.516 & 6.109 & 0.220 & 0.234 & \textbf{0.342} & \textbf{0.316} \\

MotionDiff~\cite{wei2022human}          & {\XSolidBrush}                                           & \multicolumn{1}{c|}{4} & \textbf{15.353} & 0.411 & 0.509 & 0.508 & 0.536 & 5.931 & 0.232 & 0.236 & 0.352 & 0.320

\\ \hline
\rowcolor{LightGrey} HumanMAC & {\Checkmark}                                           & \multicolumn{1}{c|}{1} & 6.301      & \textbf{0.369}     & 0.480     & 0.509     & 0.545     & \textbf{6.554}    & \textbf{0.209}     & \textbf{0.223}     & \textbf{0.342}     & 0.335     \\ 
\bottomrule
\end{tabular}
\vspace{-2.6mm}
\caption{Experimental results of quantitative results. Bolded numbers denote the state-of-the-art results. The lower is the better for all metrics except for APD. The symbol `-' indicates that the results are not reported in the baseline work.}
\vspace{-2.0mm}
\label{table:quantitative}
\end{table*}
\end{center}

\begin{figure*}
    \centering
    \includegraphics[scale=0.52]{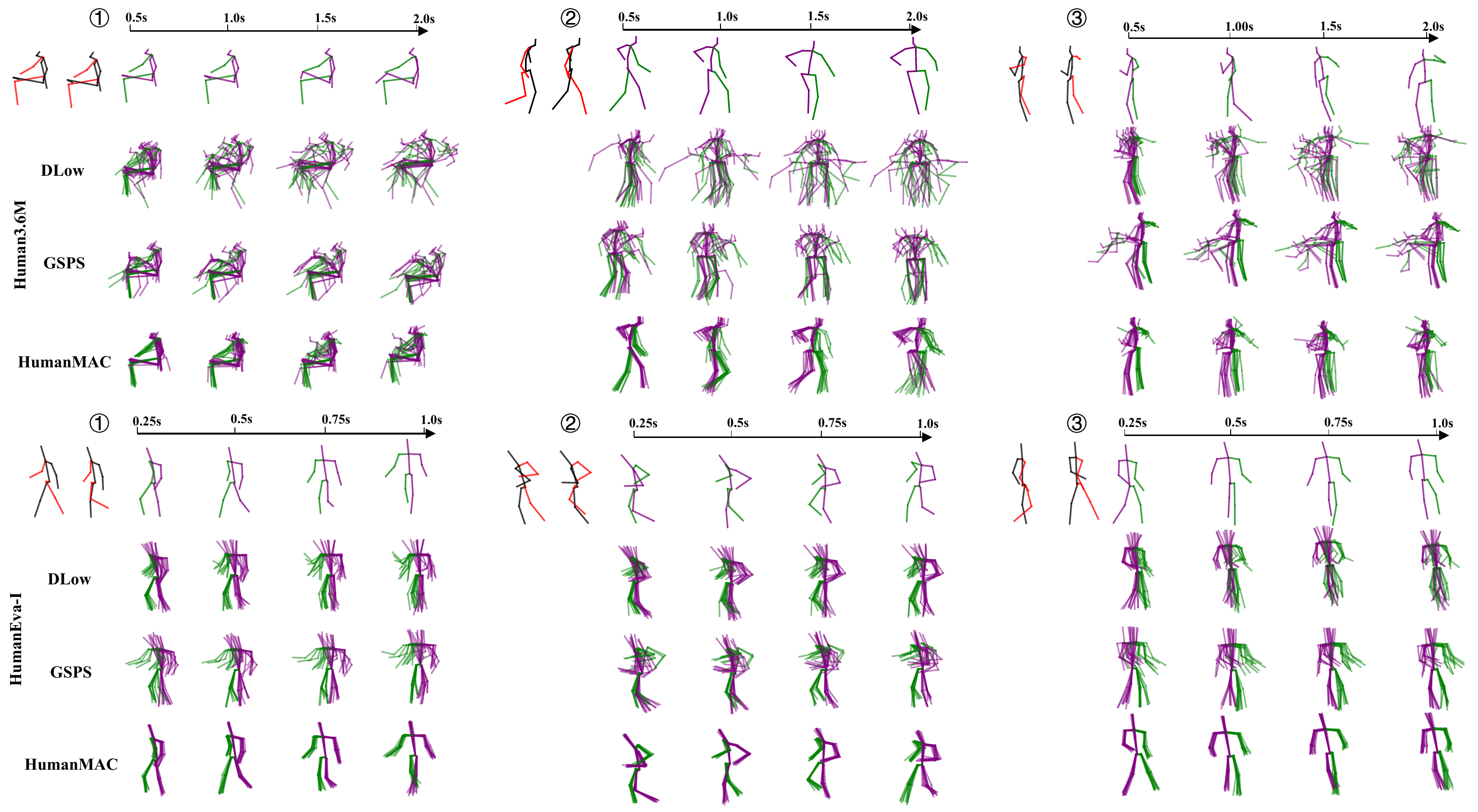}
    \vspace{-4mm}
    \caption{Visualization results. The first and the second row denote the predicted results of Human3.6M~\cite{ionescu2013human3} and HumanEva-I~\cite{sigal2010humaneva} datasets. The \textit{red-black} skeletons and \textit{green-purple} skeletons denote the observed and predicted motions respectively.}
    \vspace{-20pt}
    \label{fig:MainVis}
\end{figure*}

\vspace{-15pt}
\section{Experiments}
\vspace{-2pt}
\subsection{Experimental Setup}
\vspace{-2pt}
\myPara{Datasets.} We evaluate our model on two human motion datasets, \textit{i.e.}, Human3.6M~\cite{ionescu2013human3} and HumanEva-I~\cite{sigal2010humaneva}. For fair comparisons, we follow the same dataset partition settings with previous work. Additionally, we evaluate our method on the CMU-MoCap and AMASS dataset, shown in Section~\ref{sec:moreexp}.

\myPara{Metrics.} Following the previous work~\cite{yuan2020dlow}, we use five metrics to evaluate our model. (1) 
Average Pairwise Distance (\textit{APD}) is the L2 distance between all motion examples. It is used for measuring result diversity. (2) Average Displacement Error (\textit{ADE}) is defined as the smallest average L2 distance between the ground truth and predicted motion. It measures the accuracy of the whole sequence. (3) Final Displacement Error (\textit{FDE}) is the L2 distance between the prediction results and ground truth in the last prediction frame. (4) Multi-Modal-ADE (\textit{MMADE}) is the multi-modal version of ADE, whose ground truth future motions are grouped by similar observations. (5) Multi-Modal-FDE (\textit{MMFDE}), similarly, is the multi-modal version of FDE.

\myPara{Baselines.} In quantitative
comparison, we compare our method with several state-of-the-art works, including acLSTM~\cite{zhou2018autoconditioned}, DeLi GAN~\cite{gurumurthy2017deligan}, MT-VAE~\cite{yan2018mt}, BoM~\cite{bhattacharyya2018accurate}, DSF~\cite{yuan2019diverse}, DLow~\cite{yuan2020dlow}, GSPS~\cite{mao2021generating}, MOJO~\cite{zhang2021we}, BeLFusion~\cite{barquero2022belfusion}, DivSamp~\cite{dang2022diverse}, and MotionDiff~\cite{wei2022human}. In visualization comparison, the competitors are DLow and GSPS. 

\myPara{Implementation details.} We train HumanMAC on both datasets with a 1000-step diffusion model and sample with a 100-step DDIM~\cite{song2020denoising}. The Cosine scheduler~\cite{nichol2021improved} is exploited for variance scheduling in our model. The noise prediction network contains 8-layer TransLinear blocks for Human3.6M and 4-layer TransLinear blocks for HumanEva-I. Experiments are conducted with PyTorch~\cite{paszke2019pytorch}
and one NVIDIA Tesla A100-80GB GPU. Please refer to Appendix~\ref{sec:detail} for more details.

\vspace{-2pt}
\subsection{Comparison with the State-of-the-Arts}
\vspace{-2pt}
\myPara{Quantitative results.} Experimental results are provided in Table~\ref{table:quantitative}. As can be seen, for Human3.6M, our method achieves state-of-the-art results on both ADE and FDE metrics, which shows the plausibility of our results. Although we cannot achieve as good results as the state-of-the-art baselines in terms of diversity, as shown in Figure~\ref{fig:MainVis}, our results are more reasonable (\textit{fewer failure examples compared with baselines}). Therefore, we did not achieve the superiority of APD as an artificial indicator by introducing failure examples. For HumanEva-I, we present the competitive results on the MMFDE metric and the state-of-the-art results on other metrics. Our prediction results on HumanEva-I achieve a great trade-off between diversity and authenticity.

\myPara{Visualization results.} We present our qualitative analysis results via visualizing motion sequences in Figure~\ref{fig:MainVis}. We compare our method with DLow~\cite{yuan2020dlow} and GSPS~\cite{mao2021generating}. The first and the second row denotes three predicted results of Human3.6M~\cite{ionescu2013human3} and HumanEva-I~\cite{sigal2010humaneva} datasets respectively. In each frame, we show 10 results of the predicted motion stochastically. 

Although the DLow and GSPS methods seem more diverse, both of them generate some failure cases with large distances to ground truths, which are not reasonable motions. For instance, in the second example from Human3.6M, some generated result of DLow is floating in the air. Besides, the third generated result in Human3.6M of GSPS starts with a sudden bend and holds for a few seconds, which does not satisfy the physical constraints of the human center of gravity. We claim that the diversity exhibited by artificial metrics and the comparison of multiple predicted results are unreasonable, resulting in a large number of failure cases. By contrast, the diversity of predicted motions by our method is more reasonable than baseline methods. We  provide more empirical evidence in Appendix~\ref{sec:supp_vis_motion_predicition}. 

\subsection{Motion Switch Ability}
\label{sec:transfermain}
\vspace{-2pt}
\begin{figure*}
    \centering
    \includegraphics[scale=0.87]{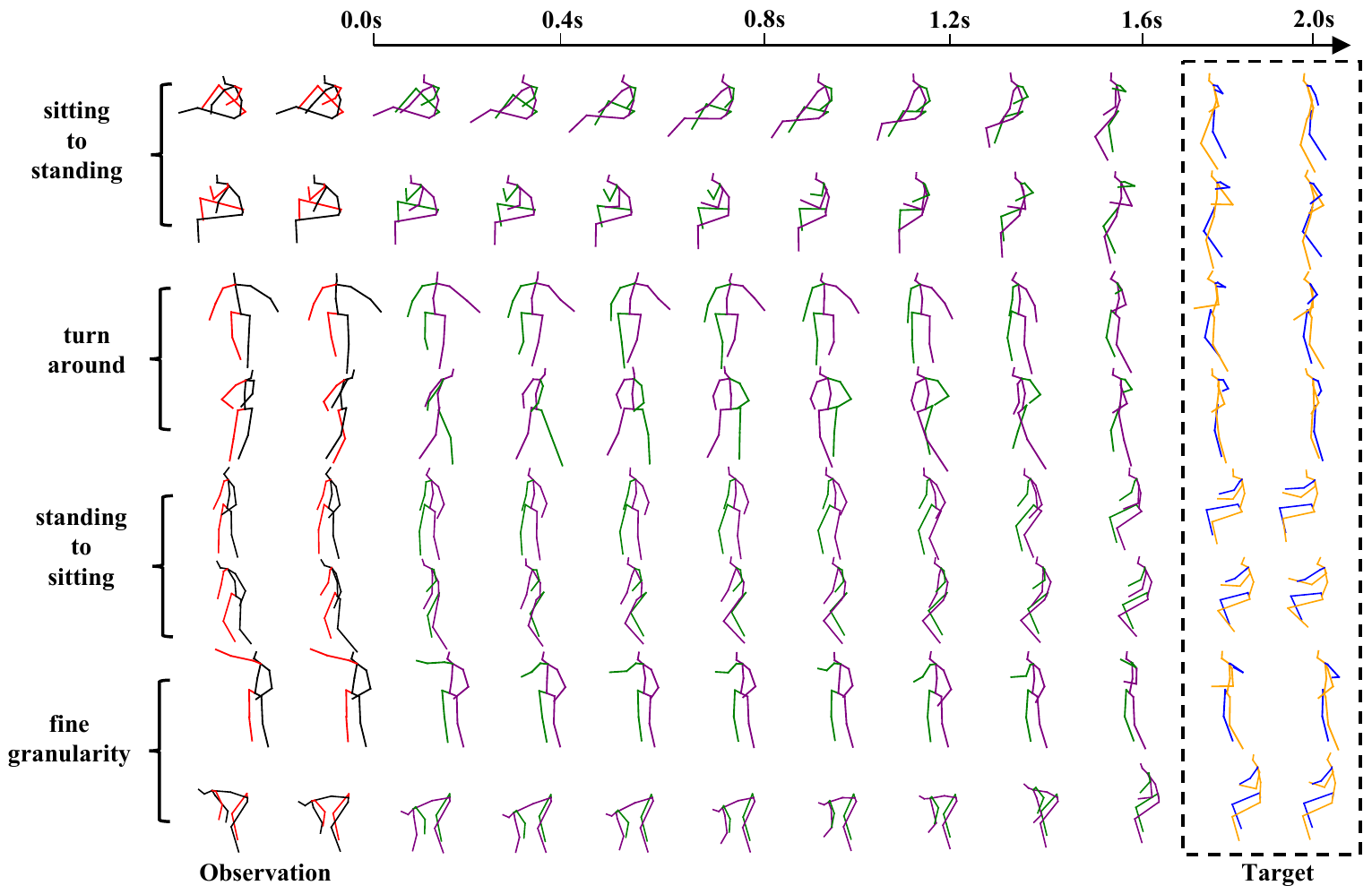}
    \vspace{-1.8mm}
    \caption{Motion switch results. Visualization of motion switch using DCT-Completion from the Human3.6M dataset. The left side shows randomly sampled motions switched to a standing pose. The right side shows various motions switched to sitting. The whole switching process is fairly smooth. The \textit{red-black} skeletons and \textit{green-purple} skeletons denote the observed and predicted motions respectively.}
    \label{fig:transfer}
    \vspace{-1.5mm}
\end{figure*}

The proposed HumanMAC enables the switch ability between different categories of motions by applying free-training DCT-Completion. The ability helps produce some motion combinations that are rare or unseen during training. 
We visualize the results of the switched motion in Figure~\ref{fig:transfer}. It shows that various observed motions switch to target motions smoothly. In the case where the directions of the human body facing to differ significantly, HumanMAC accomplishes the switch from one direction to the other smoothly and naturally. For transferring between two motions with \textit{a large distribution gap} (like from \texttt{Sitting} to \texttt{Standing}), the motion of the upper and lower bodies changes in a natural way. 
The motion switch ability shows that our model can generate sparsely distributed motion data (out-of-distribution~\cite{li2022out, xia2022moderate}, unseen, or long-tailed data). We provide more examples and analysis in Appendix~\ref{sec:transfer}.

\subsection{Part-body Controllable Prediction}
\label{sec:control}
\begin{figure*}
    \centering
    \includegraphics[scale=0.44]{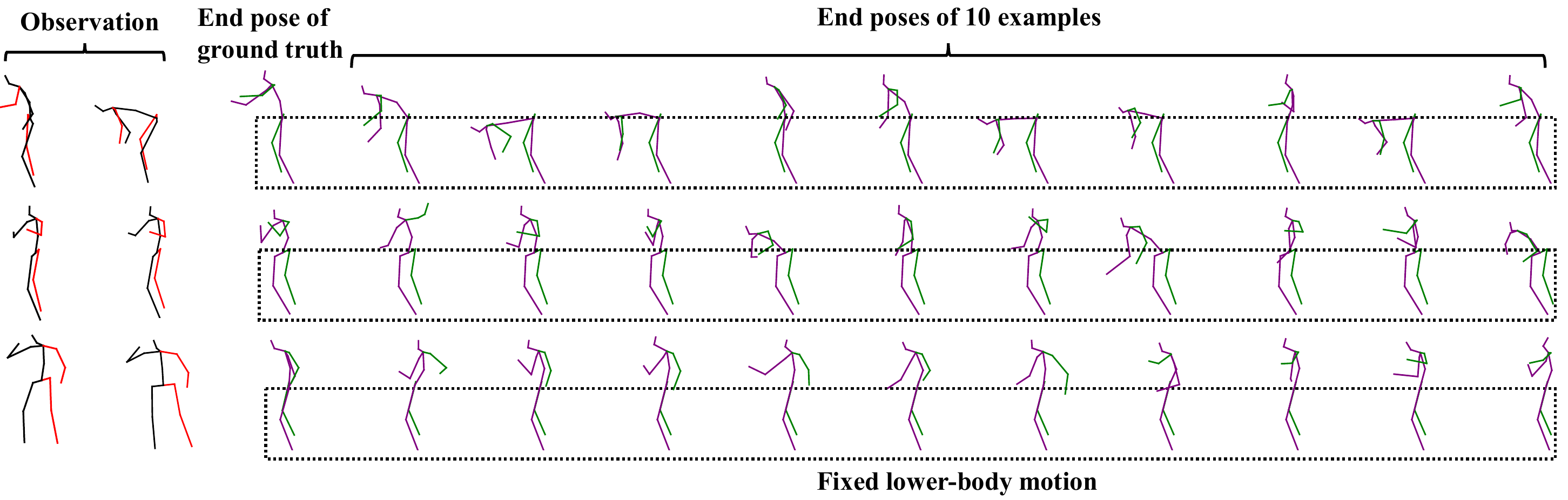}
    \vspace{-3.5mm}
    \caption{Results about controllable motion prediction, where we fix the lower body to predict the motions of the upper body. The \textit{red-black} skeletons and \textit{green-purple} skeletons denote the observed and predicted motions respectively.}
    \vspace{-12pt}
    \label{fig:control}
\end{figure*}

Due to the flexibility of the mask mechanism in the DCT-Completion algorithm (Eq.~(\ref{eq:paint})), we can complete the $F$-frame unknown part-body motion by providing the $H$-frame observation and final $F$-frame specified part-body motions. As shown in Figure~\ref{fig:control}, we fix the lower body motions and predict reasonable upper body motions. We present 10 examples of the end poses for all motions. It shows that our method can predict diverse upper body motions when keeping the limited error of the lower body. Besides, it presents some hard cases, \textit{e.g.}, large-angle turning and standing-bending transitions. The baselines GSPS and DLow only can achieve upper-/lower-body control since they need to disentangle the upper and lower body motions. Besides, with disentangled motions, these methods need to retrain the model to achieve the part-body controllable ability. However, our method does not rely on model re-training and can achieve controllable prediction of any part of the body. For more controllable part-body prediction results, please refer to Appendix~\ref{sec:part}. 

\vspace{-2pt}
\subsection{Ablation Study}
\vspace{-1pt}

\begin{table}[!t]
\centering
\small
\setlength\tabcolsep{3.3pt}
\begin{tabular}{cccc|ccc}
\toprule
\multirow{2}*{$L$} &  \multicolumn{3}{c|}{Human3.6M} & \multicolumn{3}{c}{HumanEva-I} \\
\cmidrule(lr){2-7}
& \text {APD}$\uparrow$ & \text {ADE}$\downarrow$ & \text {FDE}$\downarrow$& \text {APD}$\uparrow$ & \text {ADE}$\downarrow$ & \text {FDE}$\downarrow$ \\
\hline  5   & 6.048 & 0.388 & 0.501 & 6.444 & 0.318 & 0.340\\
        10  & \textbf{6.508} & 0.389 & 0.500 & \cellcolor{LightGrey}\textbf{6.554} & \cellcolor{LightGrey}\textbf{0.209} & \cellcolor{LightGrey}0.223\\
        20  & \cellcolor{LightGrey}6.301 & \cellcolor{LightGrey}\textbf{0.369} & \cellcolor{LightGrey}\textbf{0.480} & 5.985 & \textbf{0.209} &\textbf{0.220} \\
    \bottomrule
\end{tabular}

\caption{Experimental results of the ablation study on $L$.}
\label{table:ldim}
\end{table}

\begin{table}[!t]
\centering
\small
\setlength\tabcolsep{3pt}
\begin{tabular}{cccc|ccc}
\toprule
\multirow{2}*{\# Layer} &  \multicolumn{3}{c|}{Human3.6M} & \multicolumn{3}{c}{HumanEva-I} \\
\cmidrule(lr){2-7}
& \text {APD}$\uparrow$ & \text {ADE}$\downarrow$ & \text {FDE}$\downarrow$& \text {APD}$\uparrow$ & \text {ADE}$\downarrow$ & \text {FDE}$\downarrow$ \\
\hline  2   & 6.654 & 0.425 & 0.542 & 5.827 & 0.216 & 0.230 \\
        4   & \textbf{7.020} & 0.496 & 0.525 & \cellcolor{LightGrey}\textbf{6.554} &\cellcolor{LightGrey} 0.209 & \cellcolor{LightGrey}0.223 \\
        6   & 6.944 & 0.388 & 0.499 & 6.287 & \textbf{0.202} & \textbf{0.216} \\
        8   & \cellcolor{LightGrey}6.301 & \cellcolor{LightGrey}\textbf{0.369} & \cellcolor{LightGrey}\textbf{0.480} & 5.587 & 0.209 & 0.221 \\
        10  & 6.512 & 0.373 & 0.483 & 3.350 & 0.208 & 0.221 \\
    \bottomrule
\end{tabular}
\caption{Experimental results of the ablation study on the number of layers in the noise prediction network.}
\label{table:layers}
\end{table}

\begin{table}[!t]
\small
\centering
\setlength\tabcolsep{3pt}
\begin{tabular}{cccc|ccc}
    \toprule
\multirow{2}*{Scheduler} &  \multicolumn{3}{c|}{Human3.6M} & \multicolumn{3}{c}{HumanEva-I} \\
\cmidrule(lr){2-7}
& \text {APD}$\uparrow$ & \text {ADE}$\downarrow$ & \text {FDE}$\downarrow$& \text {APD}$\uparrow$ & \text {ADE}$\downarrow$ & \text {FDE}$\downarrow$ \\
\hline               Linear   & 6.231 & 0.379 & 0.488 & 6.320 & 0.211 & 0.233 \\
                     Sqrt     & \textbf{8.276} & 0.582 & 0.741 & \textbf{7.636} & 0.408 & 0.490 \\
\rowcolor{LightGrey} Cosine   & 6.301 & \textbf{0.369} & \textbf{0.480} & 6.554 & \textbf{0.209} & \textbf{0.223} \\
    \bottomrule
\end{tabular}
\vspace{-3.2mm}
\caption{Experimental results of the ablation study on different schedulers in the diffusion model.}
\label{table:scheduler}
\vspace{-7.7mm}
\end{table}

\begin{table}[!t]
\small
\centering
\setlength\tabcolsep{3.3pt}
\begin{tabular}{c|cccc}
    \toprule
\multirow{2}*{\# Noising steps} & \multirow{2}*{\# DDIM steps}&  \multicolumn{3}{|c}{Human3.6M} \\
 & & \multicolumn{1}{|c}{\text {APD}$\uparrow$} & \text {ADE}$\downarrow$ & \text {FDE}$\downarrow$ \\
\hline  100 &  10 & \multicolumn{1}{|c}{\textbf{6.492}} & 0.385 & 0.493  \\
 \rowcolor{LightGrey}       1000 & 100 & \multicolumn{1}{|c}{6.301} & \textbf{0.369} & \textbf{0.480}  \\
\hline
\multicolumn{2}{c}{ }& \multicolumn{3}{c}{HumanEva-I}\\
\hline  100 & 10 & \multicolumn{1}{|c}{6.199} & 0.220 & 0.244  \\
 \rowcolor{LightGrey}       1000 & 100 & \multicolumn{1}{|c}{\textbf{6.554}} & \textbf{0.209} & \textbf{0.223}  \\
    \bottomrule
\end{tabular}
\vspace{-3.2mm}
\caption{Experimental results of the ablation study on the number of diffusion steps.}
\label{table:steps}
\vspace{-10.0mm}
\end{table}

\begin{figure*}[!t]
	\centering
	\begin{subfigure}{1.0\linewidth}
		\centering
  \includegraphics[width=1.0\linewidth]{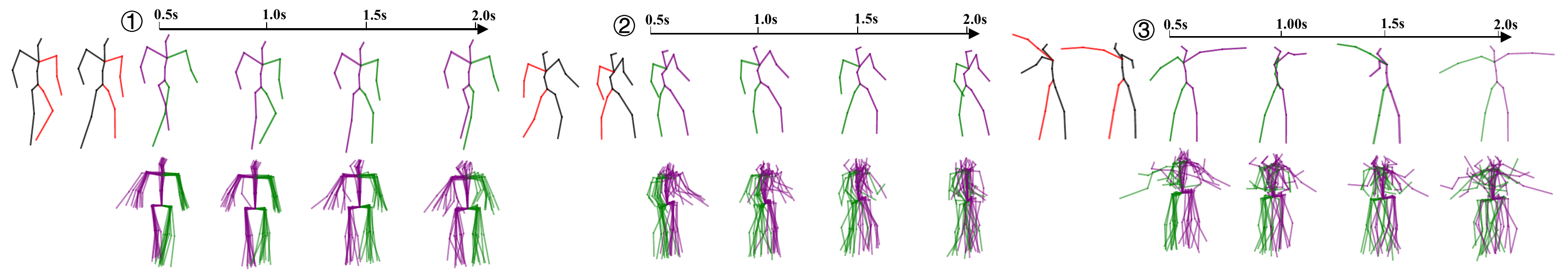}
        \caption{Motion sequence visualization. The first row is the ground truth and the second row is a sample of 10 predictions. }
		\label{amass1}
	\end{subfigure}
	\centering
	\begin{subfigure}{1.0\linewidth}
		\centering
		\includegraphics[width=1.0\linewidth]{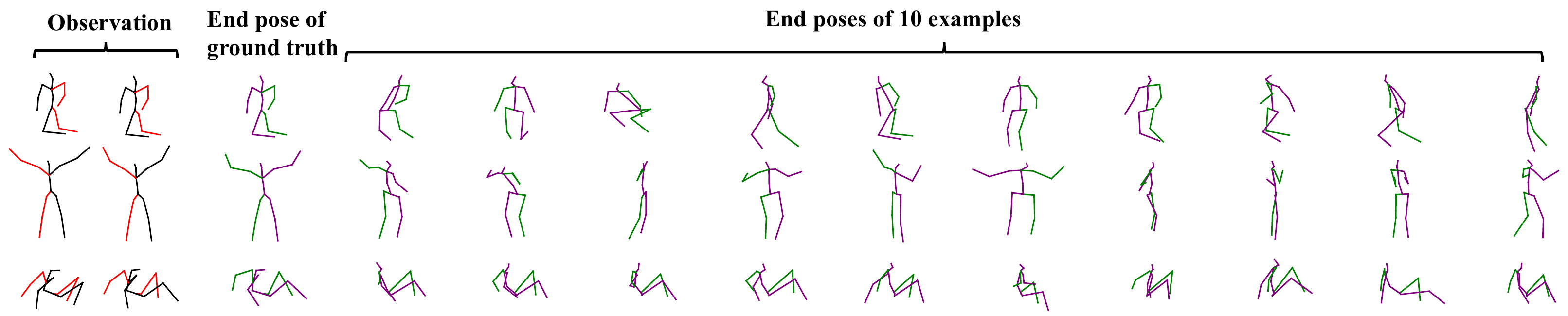}
        \caption{End pose visualization. We visualize the end pose of the prediction from three random examples.}
		\label{amass2}
	\end{subfigure}
 \caption{Visualization results of zero-shot adaption ability on the AMASS dataset. The \textit{red-black} skeletons and \textit{green-purple} skeletons denote the observed and predicted motions respectively.}
 \vspace{-10pt}
 \label{fig:amass}
\end{figure*}
We conduct comprehensive ablation studies on our methods, including (1) the value of $L$ in DCT/iDCT; (2) the design of the noise prediction network; (3) the settings of the diffusion model. We detail them as follows.
The gray shadows in tables that are highlighted in all tables indicate our choice for our method. 

\myPara{Value of $L$.}
As discussed in Section~\ref{sec:dct}, we approximate the DCT and iDCT operation by selecting the first $L$ rows of $\bm{D}$ as $\bm{D}_L$ to improve computing efficiency. In Table~\ref{table:ldim}, we show the reasonable choice of different $L$. The reasonable choices of $L$ are 10 and 20 for Human3.6M~\cite{ionescu2013human3} and HumanEva-I~\cite{sigal2010humaneva} datasets respectively.

\myPara{Design of the noise
prediction network.} In Table~\ref{table:layers}, we show the reasonable choice of the different numbers of layers in the noise
prediction network. The reasonable choices of layers are 8 and 4 for Human3.6M~\cite{ionescu2013human3} and HumanEva-I~\cite{sigal2010humaneva} respectively. We present other ablation studies on the network design in Appendix~\ref{sec:skip}, \textit{e.g.}, the influence of skip connections in the noise
prediction network.

\myPara{Settings of the diffusion model.} We ablate the impacts of both the length of the denoising steps and different diffusion variance schedulers such as ``Linear'', ``Cosine'', and ``Sqrt'' on experimental results. As shown in Table~\ref{table:scheduler}, the Cosine scheduler is the best choice for scheduling. It is because the noise cannot be increased too much in the beginning steps of the diffusion process to prevent the information from being lost too quickly. The steps of noising and sampling are set to 1000 and 100 respectively in our main results. From the quantitative results in Table~\ref{table:steps}, reducing both noising steps and sampling steps by a factor of 10 damages the authenticity of generated motions. In addition, this reduction of steps results in jerky motions in visualization results. Moreover, as shown in Tables~\ref{table:ldim},~\ref{table:layers},~\ref{table:scheduler}, and~\ref{table:steps}, the diversity and the authenticity of motions are a pair of trade-offs. Although some settings are with a higher APD metric, there are many failure cases, which we do not compare in the main results.

\vspace{-0.4em}
\subsection{Zero-shot Motion Prediction}
\vspace{-0.4em}
We present the generalization ability of the HumanMAC method via zero-shot adaption experiments. Here, we train the HumanMAC model on the Human3.6M dataset and predict motions via the given AMASS dataset~\cite{mahmood2019amass} observed motions. The experimental setting is shown in Appendix~\ref{sec:detail}. The visualization results are shown in Figure~\ref{fig:amass}. As shown in Figure~\ref{fig:amass}, our zero-shot motion prediction results on Human3.6M dataset are reasonable and diverse, which shows the good generalization ability of the HumanMAC framework. For instance, as shown in the first row of Figure~\ref{amass2}, the predicted motion \texttt{Kneeling} is vivid, which is unseen in Human3.6M. We provide more results in Appendix~\ref{sec:supp_zero_shot}.

\begin{table*}[!t]
    \centering
    \small
    \setlength{\tabcolsep}{0.3em}
    \begin{tabular}{cccccc|ccccc}
    \toprule
    \multirow{2}*{} &  \multicolumn{5}{c|}{CMU-MoCap} & \multicolumn{5}{c}{AMASS} \\[-2pt]
    \cmidrule(lr){2-11}
    \      &   {APD$\uparrow$}   & {ADE$\downarrow$}    & {FDE$\downarrow$}   & {MMADE$\downarrow$} & {MMFDE$\downarrow$}&   {APD$\uparrow$}   & {ADE$\downarrow$}    & {FDE$\downarrow$}   & {MMADE$\downarrow$} & {MMFDE$\downarrow$} \\[-1pt]
    \hline
    DLow& 15.206 & 0.516 & 0.544 & 0.544 & 0.547& \underline{13.170} & 0.590 & 0.612 & 0.618 & \underline{0.617} \\
    GSPS& 12.995 & 0.456 & 0.512 & 0.522 & 0.534 & 12.465 & 0.563 & 0.613 & 0.609 & 0.633 \\
    DivSamp& \textbf{20.473} & \underline{0.441} & \underline{0.501} & \textbf{0.508} & \underline{0.525} & \textbf{24.724} & 0.564 & 0.647 & 0.623 & 0.667 \\
    \hline
    MoDiff & \underline{19.174}  & 0.464  & 0.523  & 0.531  & 0.542  & -  & -  & -  & -  & -  \\
    BeLFusion & -  & -  & -  & -  & -  & 9.376  & \underline{0.513}  & \underline{0.560}  & \textbf{0.569} & \textbf{0.591}  \\
    \hline
    \rowcolor{LightGrey} HumanMAC & 8.021  & \textbf{0.433} & \textbf{0.467} & \underline{0.512}     & \textbf{0.505}  & 9.321 & \textbf{0.511} & \textbf{0.554} & \underline{0.593} & \textbf{0.591}   \\
    \bottomrule
    \end{tabular}
    \vspace{-0.6em}
    \caption{Evaluation on CMU-MoCap and AMASS datasets. The symbol `-' indicates that the results are not reported in the baseline work.}
    \label{tab:compare}
\end{table*}

\subsection{Evaluation on Both CMU-MoCap and AMASS Datasets}
\label{sec:moreexp}

We compare our HumanMAC framework with VAE-based and diffusion-based models on both CMU-MoCap~\cite{cmu_mocap} and AMASS\cite{mahmood2019amass} datasets. All parameters are well-tuned for all baselines. As shown in Table~\ref{tab:compare}, the HumanMAC method presents superior performance on both datasets, which shows the strong prediction ability for the mask-completion mechanism.

\subsection{Ablation Study on the DCT Design}
\label{sec:dct_ablation}

\begin{figure}[!h]
 \vspace{-1.0em}
\centering
    \begin{subfigure}{0.9\linewidth}
        \centering
        \includegraphics[width=0.92\linewidth]{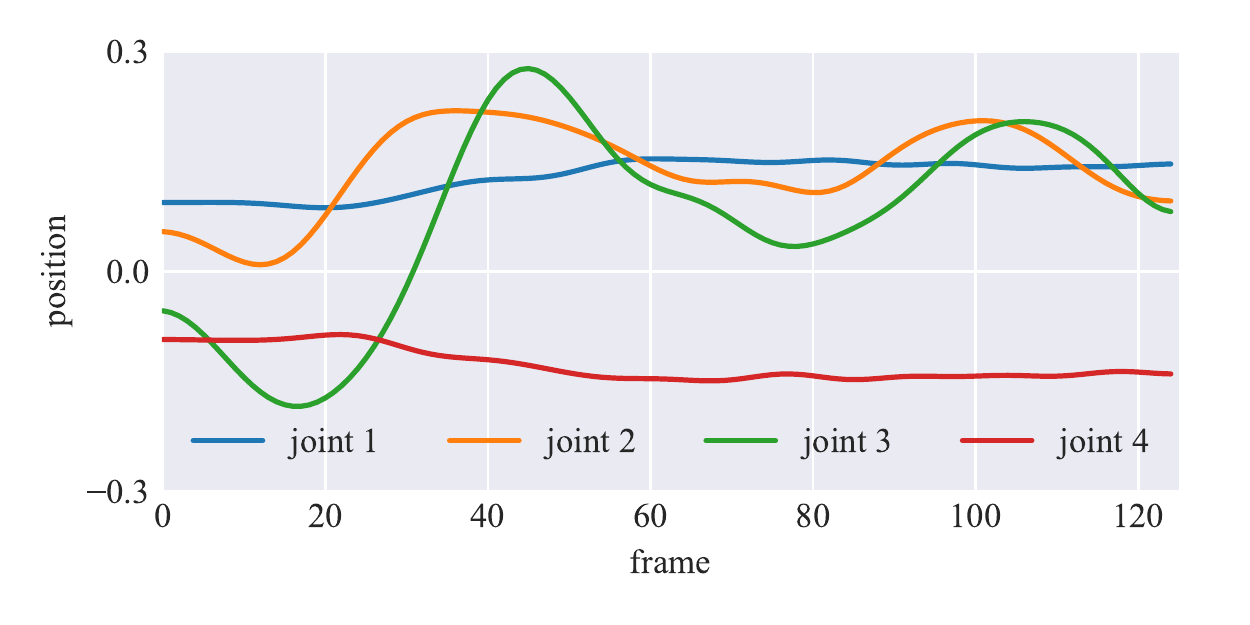}
        \vspace{-0.2em}
        \subcaption{w/ DCT.}
    \end{subfigure} 
    \begin{subfigure}{0.9\linewidth}
        \centering
        \includegraphics[width=0.92\linewidth]{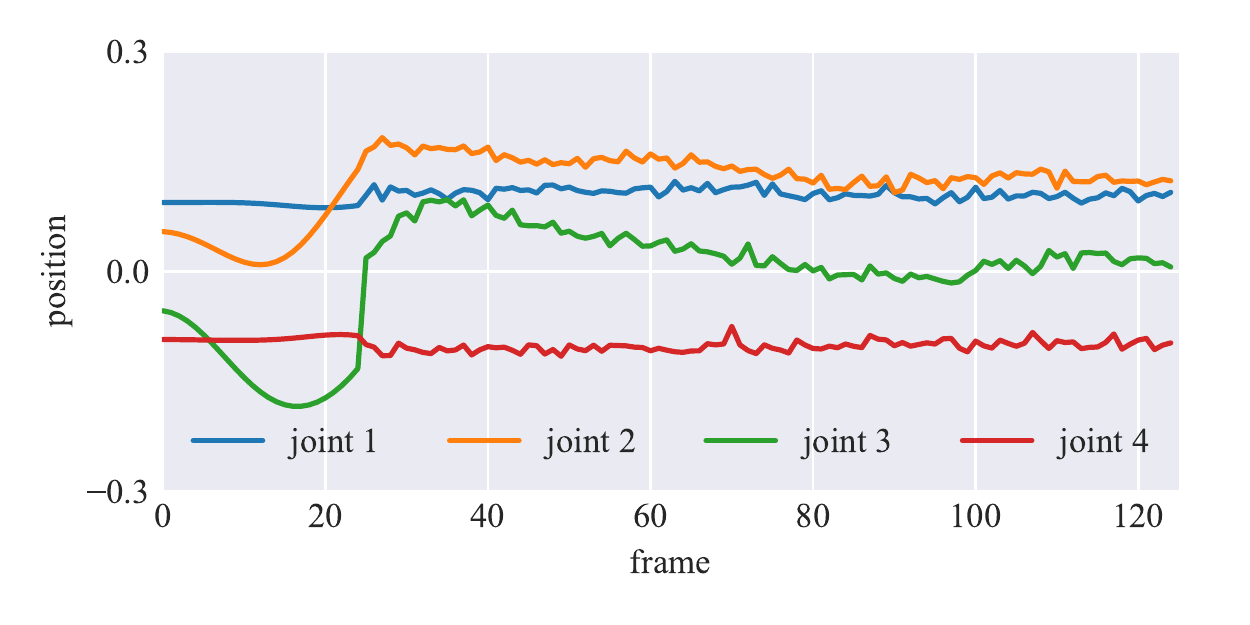}
        \vspace{-0.2em}
        \subcaption{w/o DCT.}
    \end{subfigure} 
    \vspace{-0.9em}
    \caption{Predicted motion curves on w/ or w/o DCT.}
 \label{fig:jitter}
\end{figure}

\cite{starke2022deepphase} shows the regular periodicity of motions, which inspires us to model human motion with DCT. Furthermore, we show joint motion curves (X-coordinate) of four key points in Figure~\ref{fig:jitter}, where the model w/o the DCT suffers from extreme jittering. As shown in Table~\ref{tab:jitter}, the DCT makes the prediction more accurate. In conclusion, the DCT design explicitly models the low- and high-frequency signals, making the prediction more accurate and smooth.

\begin{table}[h]
    \centering
    \small
    \begin{tabular}{cccccc}
    \toprule
    \      &   \makebox[0.01\textwidth][c]{APD$\uparrow$}   & \makebox[0.01\textwidth][c]{ADE$\downarrow$}    & \makebox[0.01\textwidth][c]{FDE$\downarrow$}   & \makebox[0.065\textwidth][c]{MMADE$\downarrow$} & \makebox[0.065\textwidth][c]{MMFDE$\downarrow$}\\
    \hline
    w/o DCT& \textbf{7.191} & 0.444 & 0.521 & 0.521 & 0.550 \\
    \hline
    \rowcolor{LightGrey} w/ DCT & 6.301  & \textbf{0.369} & \textbf{0.480}     & \textbf{0.509}     & \textbf{0.545} \\
    \bottomrule
    \end{tabular}
    \caption{Ablation on w/ or w/o DCT process (Human3.6M).}
    \label{tab:jitter}
\end{table}

\subsection{Long Time Series Prediction}

\begin{figure}[!t]
\centering
    \begin{subfigure}[b]{1\linewidth}
        \centering
        \vspace{-0.9em}
        \includegraphics[width=0.93\linewidth]{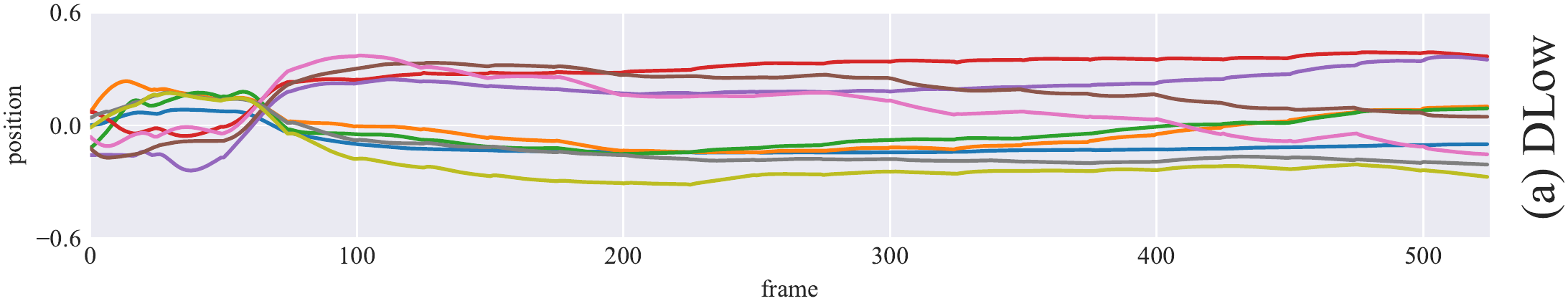}
    \vspace{-0.1em}
    \end{subfigure} 
    \begin{subfigure}{0.92\linewidth}
        \centering
        \includegraphics[width=1.0\linewidth]{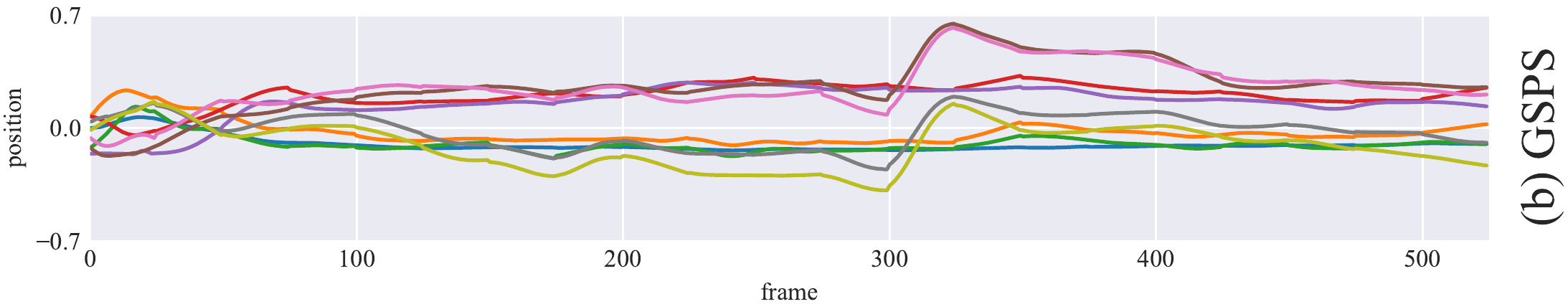}
    \vspace{-1.52em}
    \end{subfigure} 
    \vspace{-1.0em}
    \begin{subfigure}{0.92\linewidth}
        \centering
        \includegraphics[width=0.99\linewidth]{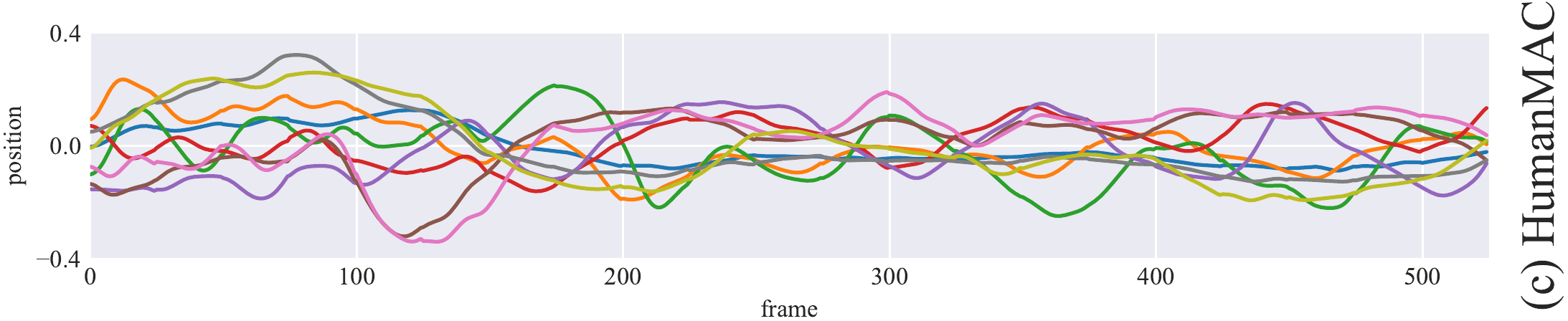}
    \vspace{-0.6em}
    \end{subfigure} 
    \caption{Joints motion curves on the X-coordinate.}
    \vspace{-1.0em}
 \label{fig:long}
\end{figure}

The long-time motion prediction can be implemented in an auto-regressive way. Technically, we can treat the final $H$-frame predictions as observations and predict further motions recurrently. In Figure~\ref{fig:long}, we present joint motion curves (X-coordinate). Baselines tend to predict \textit{over-smoothing and almost static} motions. In contrast, ours are \textit{more diverse}, which benefits from our DCT modeling.

\section{Limitation and Discussion}
HumanMAC is a diffusion-style method for the HMP task. In our diffusion model, the proposed method needs 100 steps DDIM sampling steps, which makes it limited to being a real-time system. Therefore, we will try more methods, \textit{e.g.}, DPM-solver$++$~\cite{ludpm, lu2022dpm} to reduce the sampling steps in future work. Besides, although our method is relatively simple at the present stage, we will try to further simplify the network architecture and completion algorithm. 
Moreover, we will test our method on larger-scale datasets~\cite{cai2022humman, mahmood2019amass, guo2020action2motion, guo2022generating, mandery2015kit, punnakkal2021babel, lin2023motion} for human motion prediction to build a generalized HumanMAC model and explore its zero-shot transfer ability on the HMP task.  

As analyzed in the experiment section, the previous metric for judging diversity is questionable, because of those failure cases. Therefore, in the HMP task, a more reasonable metric for diversity needs to be raised imminently. This is also what we are interested in.

\section{Conclusion}
In this paper, we target human motion prediction, which is a practical and challenging problem. We precisely reveal the issues of previous methods that work in an encoding-decoding fashion. Different from them, we propose a new learning framework for human motion prediction, which operates in a masked completion fashion. The proposed framework enjoys excellent algorithmic properties and achieves state-of-the-art performance in both quantitative and qualitative comparisons. We believe that this work can offer a new perspective to the research community of human motion prediction and inspire follow-up research. 

\section*{Acknowledgement}

We would like to thank Mr. Yu-Kun Zhou from Xidian University, and Mr. Wenhao Yang from Nanjing University for providing significant suggestions and technical support. Ling-Hao Chen is partially supported by International Digital Economy Academy (IDEA). Xiaobo Xia is supported by Australian Research Council Project DE190101473 and Google PhD Fellowship. Tongliang Liu is partially supported by the following Australian Research Council projects: FT220100318, DP220102121, LP220100527, LP220200949, and IC190100031.

\clearpage
{\small
\bibliographystyle{ieee_fullname}
\bibliography{egbib}
}

\clearpage

\appendix

\section*{Appendix of ``HumanMAC: Masked Motion Completion for Human Motion Prediction''}

\section{Noise Prediction Network TransLinear}
\label{sec:network}

\begin{figure}[h]
    \centering
    \includegraphics[scale=0.54]{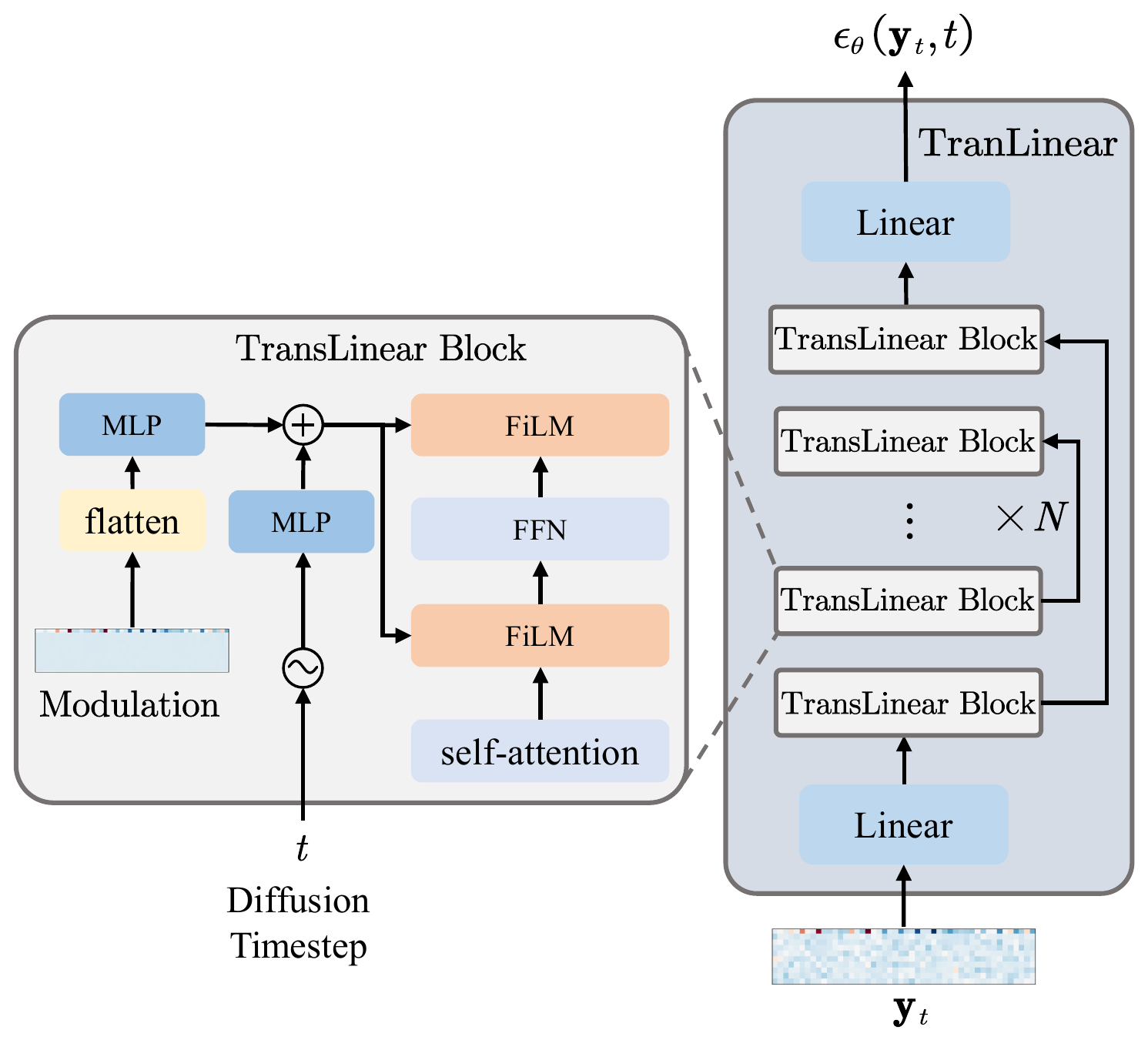}
    \caption{The architecture of the noise prediction network TransLinear, which takes the DCT spectrum $\mathbf{y}_t$ at the diffusion timestep $t$ as input. TransLinear is composed of $N$ blocks with skip connections and a linear layer. 
    }
    \label{fig:architecture}
\end{figure}

The proposed noise prediction network, \textit{i.e.}, TransLinear, is shown in Figure~\ref{fig:architecture}. The input of TransLinear is the DCT spectrum at the step $t$, noted as $\mathbf{y}_{t} \in \mathbb{R}^{L\times (H+F)}$. TransLinear has two linear layers for both input and output to map the joint's dimension. Besides, $N$ TransLinear blocks are stacked with skip connections~\cite{ronneberger2015u} in the TransLinear.  Motivated by~\cite{perez2018film}, we add two linear-based FiLM modules in the transformer encoder in each TransLinear block. To obtain temporal relationships, the FiLM module is modulated by the first $K$-frame modulating motion's DCT spectrum and the diffusion time embedding.  Since the length of the first $K$-frame modulating motions is not equal to the length of full motions, we simply to pad the last frame of the modulating motion to the full length and obtain a spectrum of the padding motion.  In summary, the TransLinear block is composed of a \textbf{Trans}former encoder and some \textbf{Linear} operations, which is a simple architecture. 

\section{Implementation Details}
\label{sec:detail}

To ensure reproducibility, we report the implementation details of HumanMAC. Source codes are public at \url{https://github.com/LinghaoChan/HumanMAC}. 

We evaluate our model on two popularly used human motion datasets, \textit{i.e.}, Human3.6M~\cite{ionescu2013human3} and HumanEva-I~\cite{sigal2010humaneva}. 
Human3.6M consists of 7 subjects performing 15 different motions, and 5 subjects (S1, S5, S6, S7, and S8) are utilized for training, while the remaining two (S9 and S11) are utilized for evaluation. We apply the original frame rate (50 Hz) and a 17-joint skeleton removing the root joint to build human motions. Our model predicts 100 frames (2s) via 25 observation frames (0.5s).
HumanEva-I comprises 3 subjects each performing 5 actions. We apply the original frame rate (60 Hz) and a 15-joint skeleton removing the root joint to build human motions. We predict 60 frames (1s) via 15 (0.25s) frames.

For both datasets, the batch size is set to 64. The model is trained for 500 epochs. The optimizer is set as Adam~\cite{kingma2014adam}. The learning rate is $3\times 10^{-4}$ with a multi-step learning rate scheduler ($\gamma=0.9$). The dropout rate is 0.2. In the DCT/iDCT operation, the number of $L$ is set to be 20 and 10 for Human3.6M and HumanEva-I respectively. For the denoising diffusion model, the variance scheduler is the Cosine scheduler~\cite{nichol2021improved} with 1000 noising steps. The DDIM sampler is set to 100 steps in the sampling stage. For the network architecture, the number of the self-attention~\cite{vaswani2017attention, zhang2023trained} head is set as 8. The latent dimension is 512. In the inference stage, the modulation ratio in the HumanEva-I dataset is set as 0.5 and 1.0 for the Human3.6M dataset.
For the motion switch ability, since the content of the observation and target have been mostly recovered in the final denoising steps of the diffusion model, we replace the final 20 steps of DCT-Completion with the vanilla denoising steps, which simplifies computation. 

For the experiments of zero-shot motion prediction, we first retarget the skeleton in the AMASS dataset to the skeleton in the Human3.6M dataset by a widely used human motion retargeting tool\footnote{\url{https://theorangeduck.com/page/deep-learning-framework-character-motion-synthesis-and-editing}}. After skeleton retargeting, we inference the AMASS motion with the model trained on the Human3.6M dataset directly. 

\section{Supplementary Visualization Results of Motion Prediction on Human3.6M and HumanEva-I}\label{sec:supp_vis_motion_predicition}

\begin{figure*}[!t]
	\centering
	\begin{subfigure}{0.9\linewidth}
		\centering
  \includegraphics[width=1.0\linewidth]{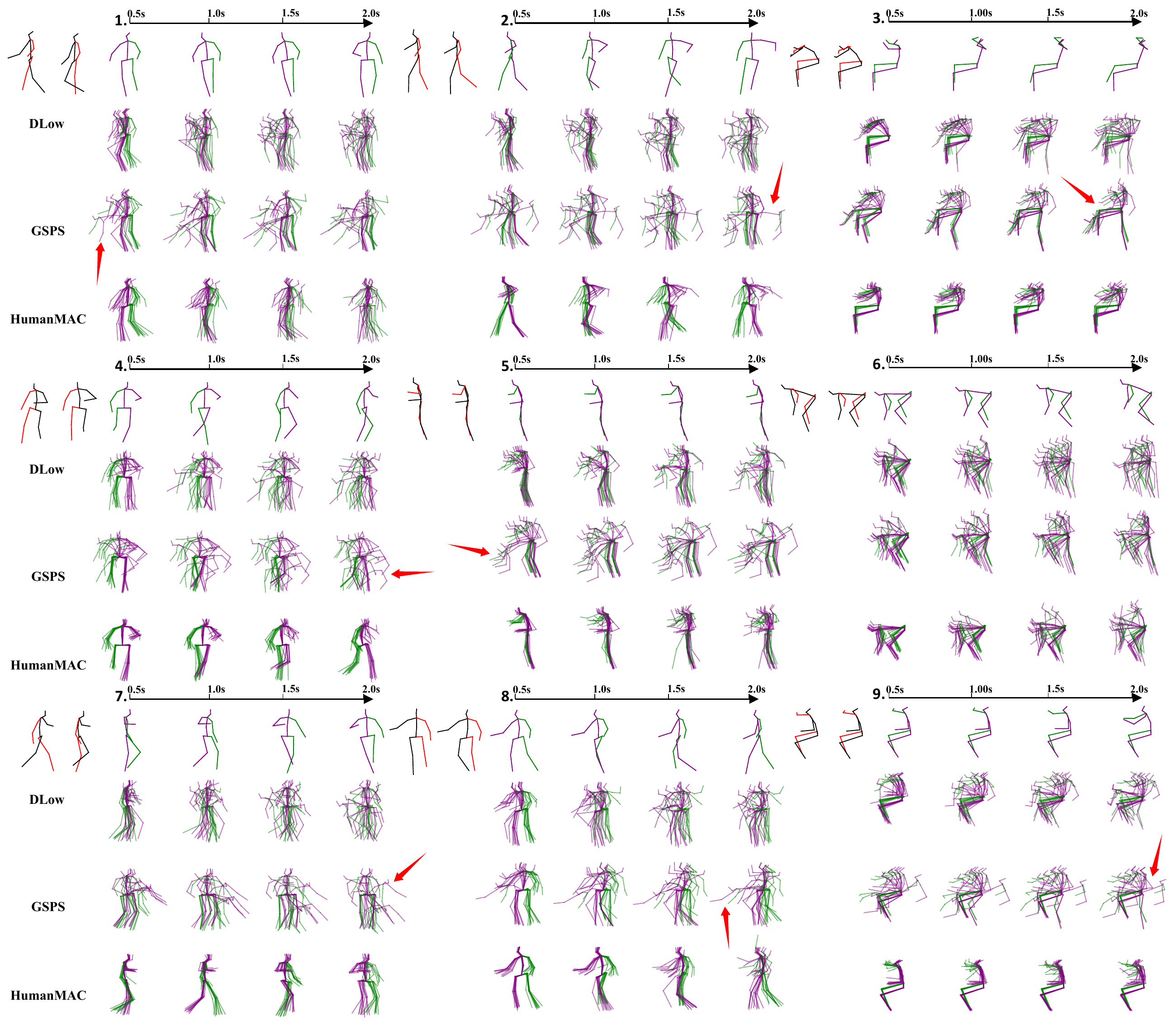}
        \caption{Comparison on motion sequences visualization.}
		\label{fig:mainvis_appendix_1}
	\end{subfigure}
	\centering
	\begin{subfigure}{0.9\linewidth}
		\centering
		\includegraphics[width=1.0\linewidth]{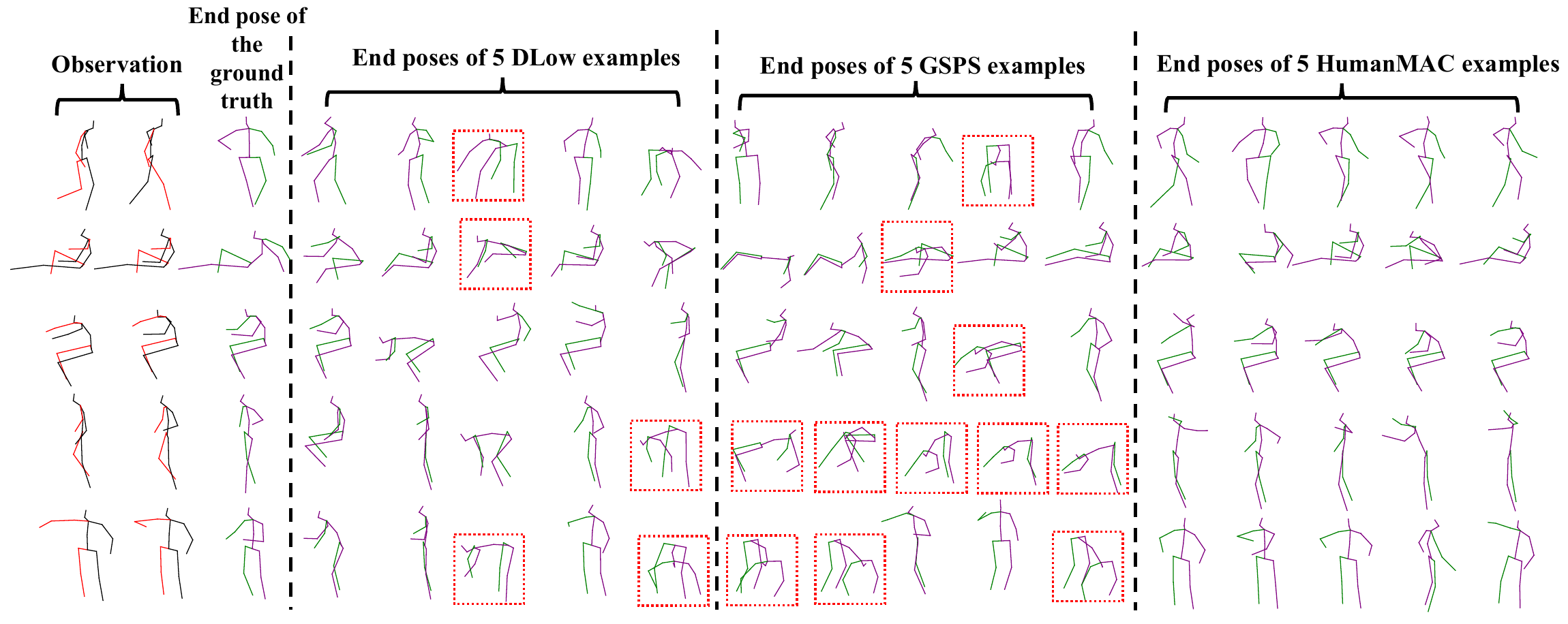}
        \caption{Comparison on end pose visualization.}
		\label{fig:mainvis_appendix_2}
	\end{subfigure}
 \caption{Visualization of motion prediction results. We present both motion sequences comparison and end-pose comparison. The \textit{red-black} skeletons and \textit{green-purple} skeletons denote the observed and predicted motions respectively.}
 \label{fig:mainvis_appendix}
\end{figure*}

We provide more empirical evidence of visualization comparison with DLow~\cite{yuan2020dlow} and GSPS~\cite{mao2021generating} in Figure~\ref{fig:mainvis_appendix}. The visualization results of motion sequences and end poses are shown in Figure~\ref{fig:mainvis_appendix_1} and Figure~\ref{fig:mainvis_appendix_2} respectively. Cases highlighted with \textbf{red arrows} in Figure~\ref{fig:mainvis_appendix_1} and end poses with \textbf{red dashed boxes} in Figure~\ref{fig:mainvis_appendix_2} are failure cases that do not satisfy the physical constraints of the human center of gravity.  By contrast, the diversity of predicted motions by our method is more reasonable than baseline methods.

\section{Motion Switch}
\label{sec:transfer}
\begin{figure*}[!t]
    \centering
    \includegraphics[scale=1.1]{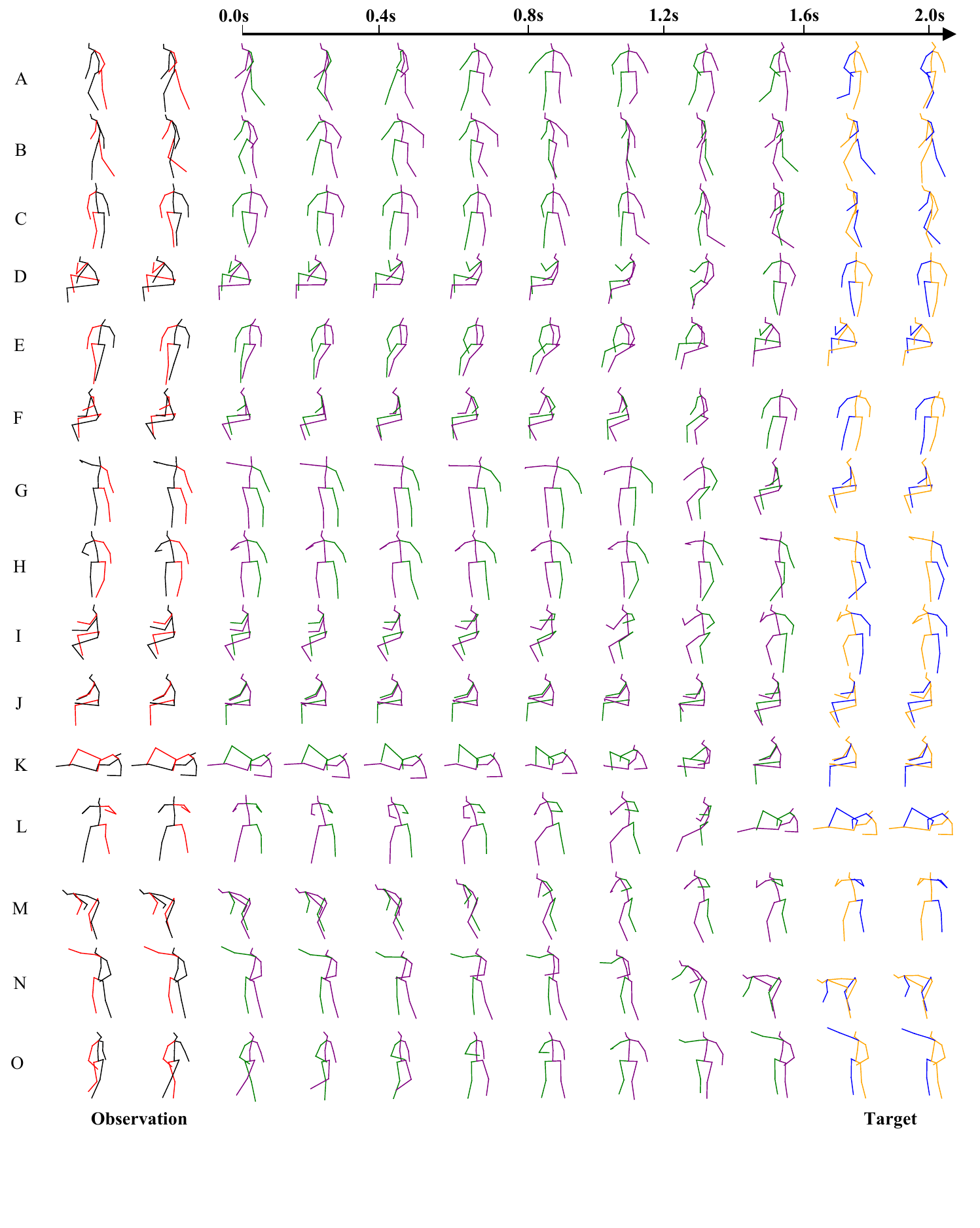}
    \vspace{-15mm}
    \caption{Additional motion switch results. Visualization of motion transfer using DCT-Completion from the Human3.6M dataset.}
    \label{fig:transfer_appendix}
    \vspace{-2mm}
\end{figure*}

In Figure~\ref{fig:transfer_appendix}, we present more results to show the motion switch ability of our method in Figure~\ref{fig:transfer_appendix}. We provide some hard cases, \textit{e.g.}, \texttt{Sitting}-\texttt{Walking} switch (example \textit{D, E, F, G, I, K}) and \texttt{Turning} (example \textit{A, C, D, E, F, I, M, O}). For transferring between two motions with \textit{a large distribution gap}, the motion of the upper and lower bodies changes in a natural way. 
We provide more animations on the project page.

\section{Part-body Controllable Prediction Results}
\label{sec:part}

\begin{figure*}
    \centering
    \includegraphics[scale=0.47]{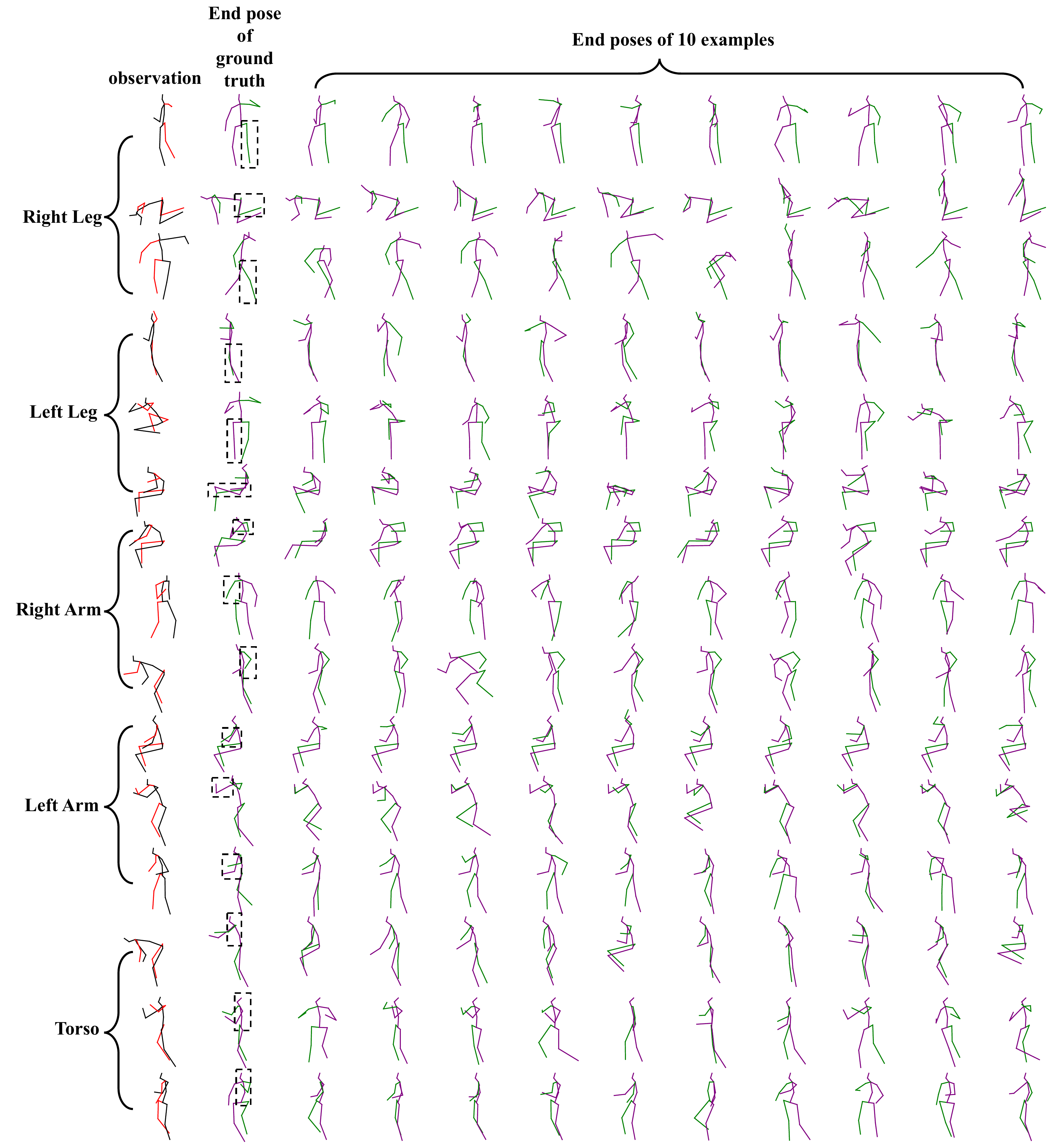}
    \caption{Flexible controllable motion prediction. We split the human skeleton into 5 parts: left leg, right leg, left arm, right arm, and torso.  In each row, we show the end poses of 10 examples of various motions when part-control certain body parts.}
    \label{fig:control_appendix}
\end{figure*}

We present more results of our part-body controllable predictions in Figure~\ref{fig:control_appendix}. As shown in Figure~\ref{fig:control_appendix}, different from previous methods, our method supports the controllability of arbitrary body parts, \textit{e.g.}, \texttt{Right Leg}, \texttt{Left Leg}, \texttt{Right Arm}, \texttt{Left Arm}, and \texttt{Torso}. This ability will facilitate controllable automatic animation synthesis.

\section{Ablation Study on Network Architecture}
\label{sec:skip}

Motivated by the U-Net~\cite{ronneberger2015u} design, a significant design in our TransLinear network is the skip connection design. As shown in Figure~\ref{table:skip}, our results show that the skip connection in the network improves the authenticity of motions. 
\vspace{-1.2em}

\begin{center}
\begin{table}[h]
\centering
\small
\setlength\tabcolsep{3.3pt}
\begin{tabular}{cccc}
    \toprule
\multirow{2}*{Mechanism} &  \multicolumn{3}{c}{Human3.6M} \\
\cmidrule(lr){2-4}
& \multicolumn{1}{c}{\text {APD}$\uparrow$} & \text {ADE}$\downarrow$ & \text {FDE}$\downarrow$ \\
\hline  w/o skip connection   & \multicolumn{1}{|c}{\textbf{6.479}} & 0.377 & 0.486 \\
        \rowcolor{LightGrey} w/ skip connection    & \multicolumn{1}{|c}{6.301} & \textbf{0.369} & \textbf{0.480}  \\
\hline
& \multicolumn{3}{c}{HumanEva-I}\\
\hline  w/o skip connection   & \multicolumn{1}{|c}{6.207} & \textbf{0.208} & 0.224  \\
        \rowcolor{LightGrey} w/ skip connection    & \multicolumn{1}{|c}{\textbf{6.554}} & 0.209 & \textbf{0.223}  \\
    \bottomrule
\end{tabular}
\vspace{-0.8em}
\caption{Ablation study on the skip connection designing.}
\vspace{-2em}
\label{table:skip}
\end{table}
\end{center}

\section{Comparison on Motion Switch and Part-body Controllable Ability}

We compare the motion switch and part-body controllable ability with baselines. As shown in Table~\ref{table:ablecompare}, our method can achieve both motion switch and part-body controllable ability. For the part-body controllable ability, DLow~\cite{yuan2020dlow} and GSPS~\cite{mao2021generating} need a specific model training stage to achieve it. In more detail, DLow needs to disentangle the human joints into two parts for training. GSPS trains two networks for the upper and lower body respectively. In our implementation, we achieve this only in the inference stage without any specific modeling. Moreover, our method supports any part-body controllable prediction.

\begin{table}[!t]
\small
\centering
\begin{tabular}{ccc}

\toprule
Method    & Switch Ability  & Part-body Controllable \\ \hline
acLSTM~\cite{zhou2018autoconditioned}      & {\XSolidBrush}           & {\XSolidBrush}           \\
DeLiGAN~\cite{gurumurthy2017deligan}      & {\XSolidBrush}           & {\XSolidBrush}           \\
MT-VAE~\cite{yan2018mt} & {\XSolidBrush} & {\XSolidBrush} \\
BoM~\cite{bhattacharyya2018accurate}        & {\XSolidBrush}           & {\XSolidBrush}           \\
DSF~\cite{yuan2019diverse}        & {\XSolidBrush}           & {\XSolidBrush}           \\
DLow$^{\star}$~\cite{yuan2020dlow}       & {\XSolidBrush}       & {\Checkmark}           \\
GSPS$^{\star}$~\cite{mao2021generating}       & {\XSolidBrush}       & {\Checkmark}           \\
MOJO~\cite{zhang2021we}       & {\XSolidBrush}           & {\XSolidBrush}           \\
BeLFusion~\cite{barquero2022belfusion}  & {\Checkmark}           & {\XSolidBrush}          \\
DivSamp~\cite{dang2022diverse}    & {\XSolidBrush}           & {\XSolidBrush}           \\
MotionDiff~\cite{wei2022human} & {\XSolidBrush}           & {\XSolidBrush}           \\ \hline 
\rowcolor{LightGrey} HumaMAC   & {\Checkmark}             & {\Checkmark} \\ \bottomrule
\end{tabular}
\caption{Comparison on motion switch and part-body controllable ability. A summary of the motion editing ability of different methods. Methods with $^{\star}$ indicate that they need specific training for achieving the part-body controllable ability.}
\label{table:ablecompare}
\end{table}

\section{Supplementary Results of Zero-shot Motion Prediction}\label{sec:supp_zero_shot}

\begin{figure*}[!h]
	\centering
	\begin{subfigure}{0.9\linewidth}
		\centering
  \includegraphics[width=1.0\linewidth]{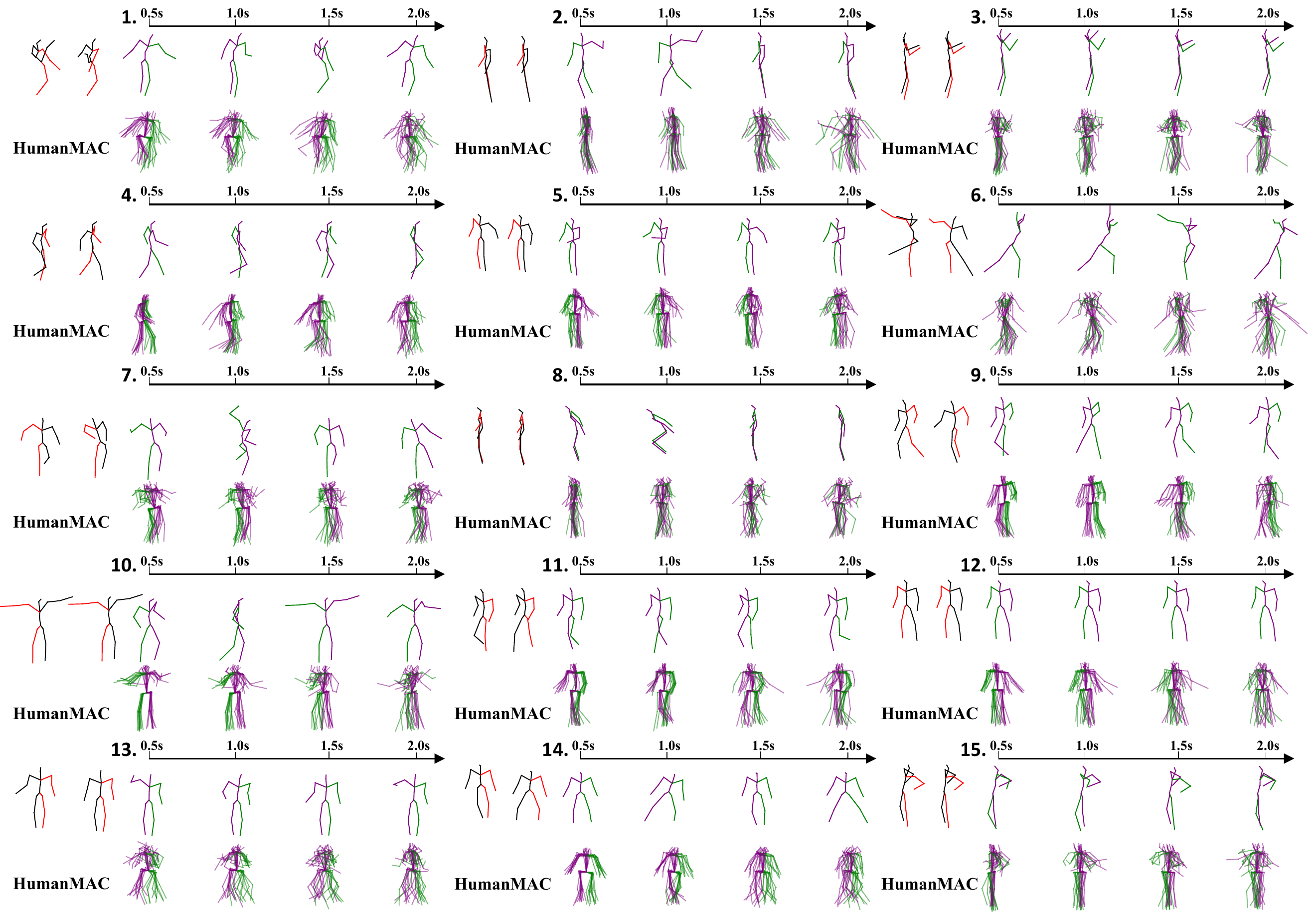}
        \caption{Motion sequences visualization. The first row is the ground truth. The second row indicates a sample of 10 predictions.}
		\label{fig:amass_appendix_1}
	\end{subfigure}
	\centering
	\begin{subfigure}{0.9\linewidth}
		\centering
		\includegraphics[width=1.0\linewidth]{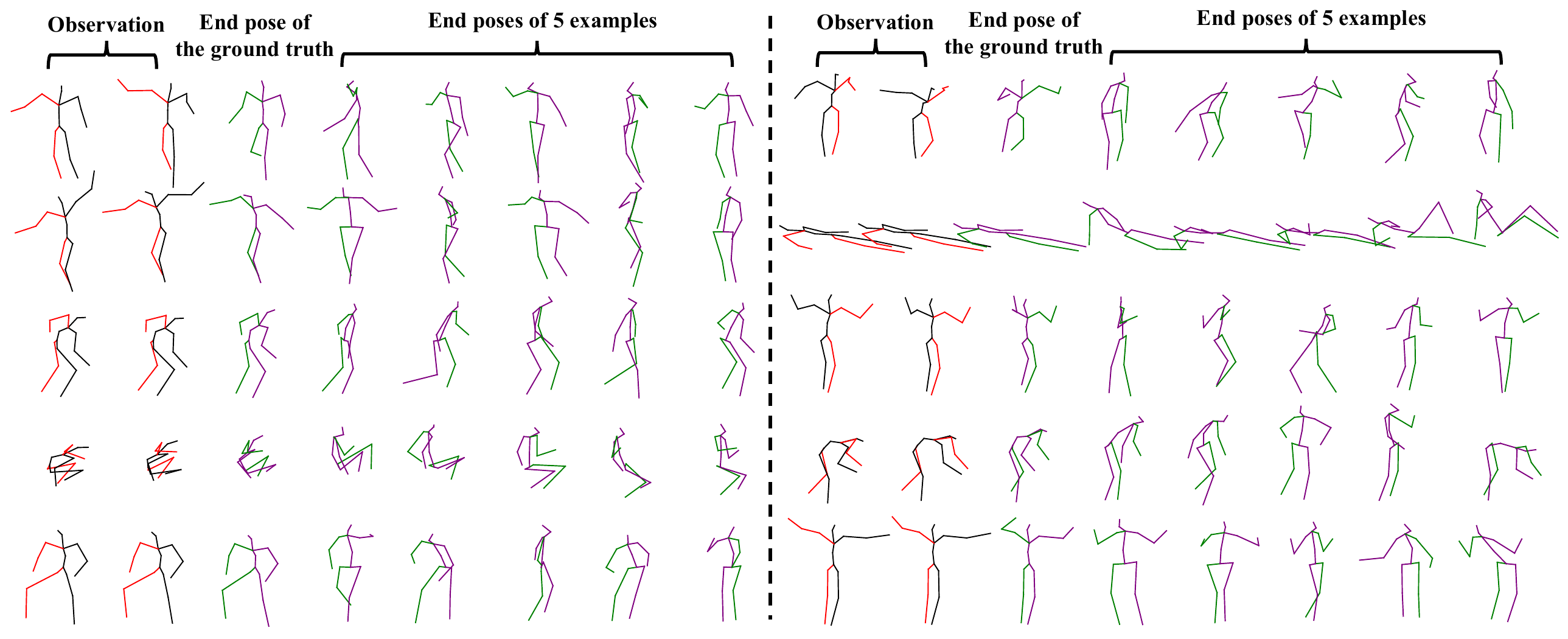}
        \caption{End pose visualization. We visualize the end pose of the prediction from three random examples.}
		\label{fig:amass_appendix_2}
	\end{subfigure}
 \caption{Visualization results of zero-shot adaption ability on the AMASS dataset. The \textit{red-black} skeletons and \textit{green-purple} skeletons denote the observed and predicted motions respectively.}
 \label{fig:amass_appendix}
\end{figure*}

We provide more empirical evidence of zero-shot motion prediction results in Figure~\ref{fig:amass_appendix}. The visualization results of predicted motion sequences and end poses are shown in Figure~\ref{fig:amass_appendix_1} and Figure~\ref{fig:amass_appendix_2} respectively. As the results shown in Figure~\ref{fig:amass_appendix}, our method can predict some motions not seen in the Human3.6M dataset, such as \textit{opening arms exaggeratedly} and \textit{kicking sharply}. See more vivid predicted motions in the supplementary video.

\section{Engineering Optimization for Evaluation}
\label{sec:engineering}

\begin{figure}[!h]
    \centering
    \includegraphics[scale=0.3]{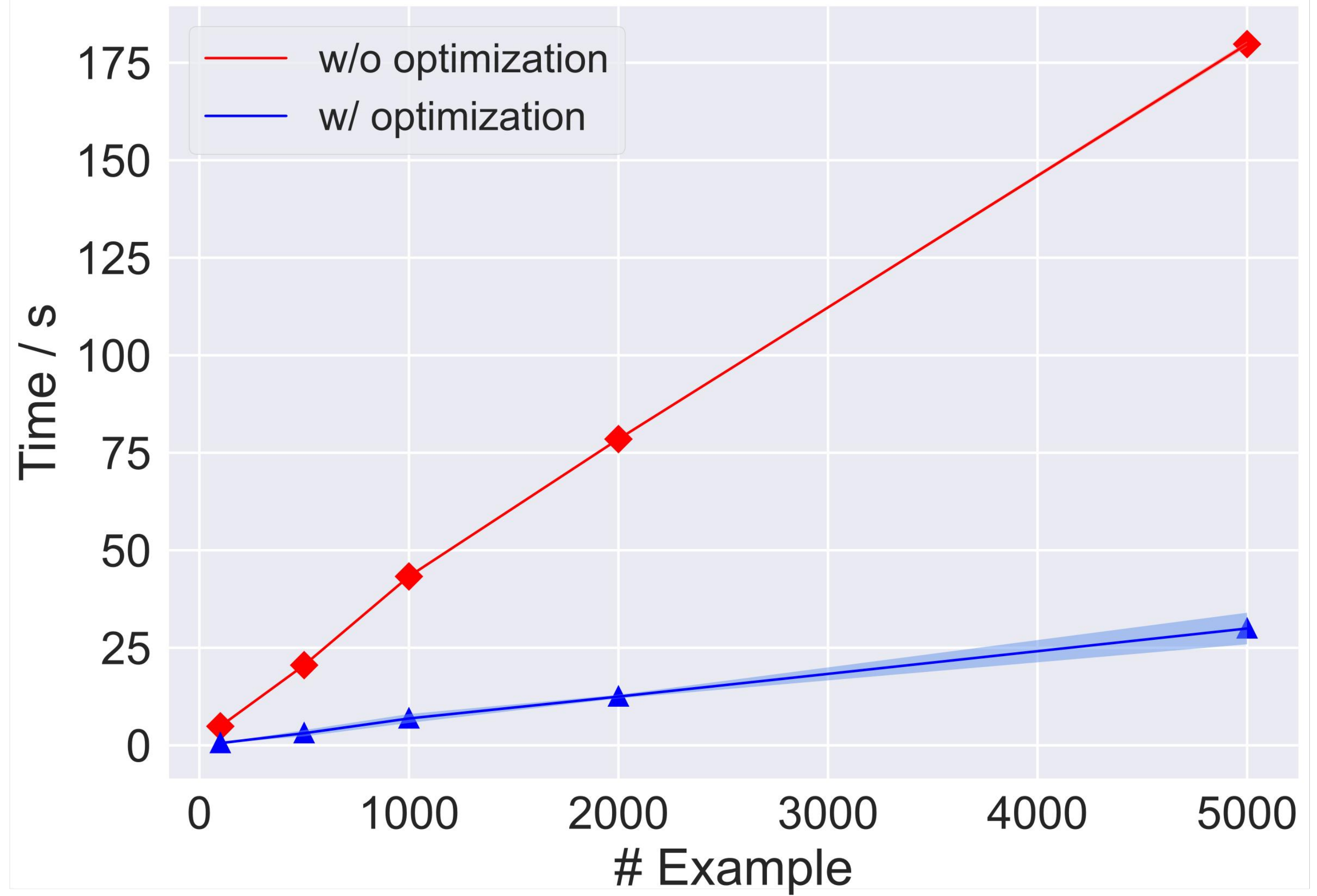}
    \caption{Engineering optimization simulation. We show the results without~(w/o) and with (w/) our engineering optimization for comparison. Our implementation improves the speed significantly. }
    \label{fig:eng}
\end{figure}

To provide more convenience for the research community, we optimize the evaluation process from an engineering aspect. The optimization consists of two aspects: (1) parallelized model inference; (2) parallelized metric calculation. We implement the evaluation on both parallelized model inference and parallelized metric calculation. For the parallelized model inference, we parallelize the serialized inference process over multiple examples in~\cite{yuan2020dlow}. We present the simulation results of the parallelized metric calculation in Table~\ref{table:eng}. The results show that our method has $\sim 6 \times$ speed up than the previous implementation\footnote{\url{https://github.com/Khrylx/DLow}}. For intuitive comparison, the comparison of the simulation is also shown in Figure~\ref{fig:eng}. After the engineering optimization, the overall (both parallelized model inference and parallelized metric calculation) speedup is $\sim$ 1k $\times$ than the previous implementation. The overall optimization speedup is shown in Table~\ref{table:overallspeedup}. For engineering optimization, we perform the experiment of engineering optimization on a machine with 30 GB memory, 32 CPU cores, and one NVIDIA Tesla A5000 GPU. For more details, please refer to \url{https://github.com/LinghaoChan/HumanMAC}. 

\begin{table}[!h]
\centering
\small
\setlength\tabcolsep{3pt}
\begin{tabular}{c|ccc}
\toprule
\# Examples & w/o optimization       & w/ optimization       & Speedup \\
\hline
100  & 4.87$\pm$0.05    & \textbf{0.56$\pm$0.05}     & $\uparrow$ 869.6\%    \\
500  & 20.55$\pm$0.21   & \textbf{3.10$\pm$0.81}     & $\uparrow$ 662.9\%    \\
1000 & 43.29$\pm$0.25   & \textbf{6.88$\pm$1.14}     & $\uparrow$ 672.8\%    \\
2000 & 78.51$\pm$0.16   & \textbf{12.49$\pm$0.48}    & $\uparrow$ 628.6\%    \\
5000 & 179.78$\pm$0.69  & \textbf{29.95$\pm$4.04}    & $\uparrow$ 600.6\%    \\
\bottomrule
\end{tabular}
\vspace{-2.6mm}
\caption{Engineering optimization simulation. We randomly generate a certain number of examples for evaluation. We show the results without~(w/o) and with (w/) our engineering optimization for comparison.}
\label{table:eng}
\end{table}

\begin{table}[!h]
\centering
\small
\setlength\tabcolsep{3pt}
\begin{tabular}{c|ccc}
\toprule
Method & w/o optimization       & w/ optimization       & Speedup \\
\hline
DLow~\cite{yuan2020dlow}  & $\sim$13 h    & \cellcolor{LightGrey}$\sim$ \textbf{52 s}     & $\sim$1k$\times$    \\
Ours  & -   & \cellcolor{LightGrey}$\sim$\textbf{16 mins}  & $ \diamond $    \\
\bottomrule
\end{tabular}
\caption{Overall evaluation optimization comparison. The symbol `-' indicates the computation time is longer than 1 day. The symbol `$\diamond$' means that the speed improvement is significant.}
\label{table:overallspeedup}
\end{table}

\end{document}